\documentclass{article} 
\usepackage{iclr2027_conference,times}

\usepackage{amsmath,amsfonts,bm}

\def\eqref#1{equation~\ref{#1}}

\def\1{\bm{1}}

\DeclareMathAlphabet{\mathsfit}{\encodingdefault}{\sfdefault}{m}{sl}
\SetMathAlphabet{\mathsfit}{bold}{\encodingdefault}{\sfdefault}{bx}{n}

\usepackage{hyperref}
\usepackage{url}

\usepackage{graphicx}
\usepackage{amssymb}
\usepackage{booktabs}
\usepackage{multirow}

\definecolor{iccvblue}{rgb}{0.21,0.49,0.74}

\hypersetup{
    colorlinks=true,
    linkcolor=red,
    citecolor=iccvblue,
    urlcolor=iccvblue
}

\definecolor{lightgreen}{RGB}{190,240,190}
\definecolor{lightyellow}{RGB}{255,245,150}

\title{
ArticuTable: Generating Instance-Level \\ \mbox{Interactive}
Rigid-Articulated 3D Tabletop \\Scenes from a Single Image
}

\author{
\href{https://johnnyboy1013.github.io/}{\textcolor{black}{Kai Lv}}\textsuperscript{*,1} \quad
\href{https://bryceyin13.github.io/}{\textcolor{black}{Yibo Yin}}\textsuperscript{*,2} \quad
\href{https://shawnricardo.github.io/}{\textcolor{black}{Lijun Guo}}\textsuperscript{1} \quad
\href{https://hengfan2010.github.io/}{\textcolor{black}{Heng Fan}}\textsuperscript{3} \quad
\href{https://zhangkaihao.github.io/}{\textcolor{black}{Kaihao Zhang}}\textsuperscript{4} \quad
\href{https://xingpingdong.github.io/}{\textcolor{black}{Xingping Dong}}\textsuperscript{\ensuremath{\dagger},1}
\\
\textsuperscript{1}School of Computer Science, Wuhan University \\
\textsuperscript{2}École Polytechnique Fédérale de Lausanne (EPFL), Switzerland \\
\textsuperscript{3}Department of Computer Science and Engineering, University of North Texas \\
\textsuperscript{4}Australian National University \\
\texttt{kailv1013@whu.edu.cn, yibo.yin@epfl.ch, xingpingdong@whu.edu.cn}\\
\textbf{Code: }
\href{https://github.com/bryceyin13/articutable-code}
{\texttt{github.com/bryceyin13/articutable-code}}
}

\iclrfinalcopy 
\begin{document}

\maketitle

\begingroup
\renewcommand{\thefootnote}{\fnsymbol{footnote}}
\footnotetext[1]{Equal contribution.}
\footnotetext[2]{Corresponding author.}
\endgroup

\begin{figure}[h]
    \centering
    \includegraphics[width=\linewidth]{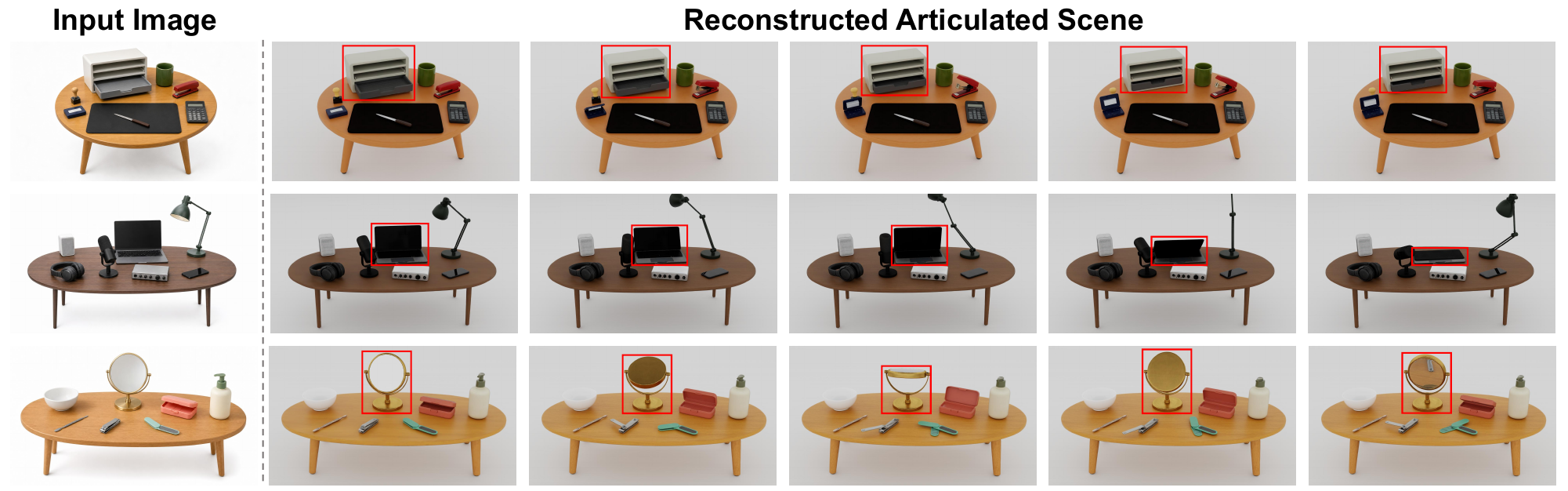}
    \caption{\textbf{Single-image reconstruction of articulated tabletop scenes.} Given one RGB image (left), ArticuTable reconstructs an input-view-consistent 3D scene containing rigid and articulated objects. Renderings of the same scene under different joint configurations (right) demonstrate part-level articulation while preserving the arrangement of object instances.}
    \label{fig:teaser}
\end{figure}

\begin{abstract}
Embodied agents benefit from 3D environments that combine visual fidelity to real-world observations with physical interactivity. Existing single-image tabletop reconstruction methods recover plausible scene geometry but typically represent objects as monolithic rigid bodies, limiting interaction to whole-object rigid motion and precluding executable part-level articulation. Meanwhile, recovering a scene layout consistent with the input view remains challenging because a single observation may admit multiple plausible pose--scale configurations. We present \textbf{ArticuTable}, a single-image 3D tabletop reconstruction framework that recovers both executable part-level articulation and an input-view-consistent scene layout. For object modeling, we introduce generation-robust articulation modeling (\textbf{GRAM}), which combines joint fitting guided by a multimodal large language model with semantic state reasoning to recover reliable joint parameters and valid motion ranges from imperfect monolithic proxy meshes, thereby converting them into executable articulated assets.
For scene layout, we introduce progressive semantic-geometric scene registration (\textbf{PSGSR}), which progressively narrows the pose--scale search space under complementary metric, planar, and input-view constraints and resolves orientation ambiguity through structure-aware semantic correspondences, yielding a scene layout consistent with the input view. We further contribute \textbf{ArticuTable-100}, a curated collection of 100 simulation-ready tabletop scenes. Extensive evaluation, including a user study, demonstrates strong performance across visual fidelity, input-view consistency, articulation quality, physical plausibility, and simulation readiness.
\end{abstract}

\section{Introduction}

Rapid advances in embodied intelligence have increased demand for large-scale collections of 3D scenes. Such scenes should faithfully reproduce object appearance and spatial layout while encoding the physical and kinematic properties required for interaction~\citep{yang2024physcene,nasiriany2024robocasa,lee2025dynscene,pfaff2025steerable,xia2026sage,li2026spread}. Tabletop scenes are particularly important because they concentrate diverse object instances and dense spatial relationships within compact workspaces, making them natural testbeds for fine-grained manipulation~\citep{wang2026tabletopgen,luo2026stable}. However, manually authoring visually faithful and physically interactive tabletop scenes is labor-intensive and difficult to scale. Automatically reconstructing such scenes from readily available monocular RGB images therefore offers a practical and scalable alternative. Recent single-image 3D scene reconstruction methods have made substantial progress in decomposing scenes into object instances, reconstructing per-instance geometry, and recovering their spatial arrangements~\citep{huang2025midi,yao2025cast,wang2026tabletopgen}.

A prevalent paradigm in these works is instance-wise reconstruction and registration~\citep{gumeli2022roca,yao2025cast,wang2026tabletopgen}. Under this paradigm, each object is independently reconstructed or retrieved as a 3D asset and then registered to a shared scene coordinate frame by estimating its pose and scale from the input view. Although this factorization provides detailed object geometry and appearance, the resulting assets typically remain monolithic surface meshes. Consequently, reconstructed scenes support object-level rigid-body interaction but lack physically meaningful part-level motion. A natural solution is to apply existing articulation modeling methods to the reconstructed assets. However, many existing methods are trained and quantitatively evaluated on curated articulated-object datasets whose assets have well-defined part structures and articulation annotations~\citep{xiang2020sapien,li2026particulate}. Image-generated proxy meshes differ substantially: part boundaries may be fused, contact geometry may be distorted, and internal structures may be incomplete. These artifacts can make joint parameters estimated from local geometry unreliable and cause motion ranges inferred from learned priors or geometric constraints to yield implausible articulated states. Consequently, directly applying existing articulation modeling methods to generated meshes may fail to produce reliable executable articulated assets. Scene registration presents a separate challenge. A single input view may admit multiple plausible pose--scale configurations, some of which attain comparable alignment scores. Existing methods commonly recover scene layouts by directly optimizing object poses and scales against the input view, resulting in a non-convex objective with competing basins of attraction. The solution therefore depends strongly on initialization, and an unfavorable initial estimate may lead local refinement to a geometrically incorrect arrangement that is inconsistent with the input view.

To address these challenges, we present \textbf{ArticuTable}, a framework for reconstructing simulation-ready tabletop scenes containing both rigid and articulated objects from a single RGB image. As illustrated in Fig.~\ref{fig:teaser}, ArticuTable preserves the observed instance arrangement while enabling executable part-level articulation; the overall framework is shown in Fig.~\ref{fig:pipeline}. For object modeling, we introduce generation-robust articulation modeling (GRAM). GRAM first decomposes a monolithic proxy mesh into geometric primitives and organizes them into functional links and a rooted kinematic tree. To make joint estimation robust to distorted proxy geometry, MLLM-guided joint fitting first constructs joint-type-specific geometric hypotheses and then uses a multimodal large language model (MLLM) with global multimodal and structural evidence to select or correct joint axes and pivots. Meanwhile, semantic state reasoning combines functional semantic evidence with geometric feasibility to infer valid motion ranges, reducing its reliance on geometric or collision constraints derived from imperfect meshes. The resulting kinematic structure, joint parameters, and motion ranges are exported as an executable URDF asset. For scene layout recovery, we introduce progressive semantic-geometric scene registration (PSGSR). Its progressive multi-constraint registration (PMCR) progressively restricts the pose--scale search space: point-cloud alignment first provides a coarse metric estimate, top-view registration then refines the planar layout, and input-view registration finally improves consistency with the original observation. Each stage is initialized by the preceding estimate, reducing the risk that subsequent local refinement enters an incorrect basin of attraction. Structure-aware semantic pose selection (SASPS) ranks yaw candidates in both stages using structure-weighted bidirectional semantic correspondences and initializes differentiable refinement with the selected pose. Efficient 2D top-view rotations reduce input-view search to residual symmetric hypotheses, enabling PSGSR to recover an input-view-consistent layout.

We further construct \textbf{ArticuTable-100}, a curated collection of 100 simulation-ready tabletop scenes containing visual meshes, collision geometry, and executable articulated assets. Quantitative comparisons, simulation-based evaluation, and a user study assess the reconstructed scenes in terms of visual fidelity, input-view consistency, articulation quality, and physical plausibility.

Our main contributions are summarized as follows:
\begin{itemize}
    \item We present \textbf{ArticuTable}, a framework that reconstructs simulation-ready tabletop scenes from a single RGB image while simultaneously enabling executable part-level articulation and preserving scene-layout consistency with the input view.

    \item We introduce \textbf{GRAM}, which combines MLLM-guided joint fitting with semantic state reasoning to address unreliable contact-derived joint estimates and mesh-induced motion-range errors, enabling executable URDF asset generation from imperfect monolithic meshes without articulation-specific training or fine-tuning.

    \item We introduce \textbf{PSGSR}, which integrates progressive geometric and visual constraints with structure-aware semantic pose reasoning to resolve pose ambiguity and progressively recover an input-view-consistent scene layout.

    \item We contribute \textbf{ArticuTable-100}, a curated collection of 100 simulation-ready tabletop scenes containing visual meshes, collision geometry, and executable articulated assets.
\end{itemize}

\section{Related Work}

\textbf{Single-Image 3D Scene Reconstruction.}
Single-image 3D scene reconstruction has progressed from holistic indoor modeling toward object-centric scene assembly for embodied interaction. Total3DUnderstanding~\citep{Nie_2020_Total3} jointly estimates room layout, camera and object poses, and object meshes, while Text2Room~\citep{Hollein_2023_Text2} synthesizes textured room-scale geometry. Feed-forward methods such as DUSt3R~\citep{Wang_2024_DUSt3R} and VGGT~\citep{Wang_2025_VGGT} directly estimate camera parameters and dense geometry. For robotic applications, Digital Cousins~\citep{Dai_2025_Auto} constructs interactive scene variations that preserve geometric and semantic affordances. More closely related to our tabletop setting, TabletopGen~\citep{wang2026tabletopgen} generates per-object assets and assembles collision-free scenes through differentiable rotation optimization and top-view spatial alignment.

\textbf{Foundation Models for Object-Centric Reconstruction.}
Foundation models connect semantic interpretation with geometric synthesis in object-centric reconstruction. Large multimodal models reason about object functions and spatial relationships~\citep{singh2026openaigpt5card,Liu_2023_Visual}, while language priors support compositional scene-layout generation~\citep{feng2023layoutgpt}. Promptable segmentation grounds these cues into object regions in images~\citep{Kirillov_2023_Segment,carion2026sam} and parts in native 3D geometry~\citep{ma2025p3sam}. Image-conditioned generators then lift isolated object observations into detailed open-domain 3D assets~\citep{hunyuan3d22025tencent,Xiang_2025_Structured,Xiang_2026_Native}.

\textbf{Articulated Object Modeling.}
For interactive tabletop scenes, reconstructed static geometry must be endowed with part structures, kinematic relationships, and joint constraints. Given observations before and after interaction or across articulation states, motion can be inferred from geometric changes~\citep{Jiang_2022_Ditto,Liu_2023_PARIS}. Image-based methods use learned priors, retrieval, or vision-language reasoning~\citep{jiang2022opd,chen2024urdformer,liu2025singapo,zhao2025real2code,le2025articulateanything}. Static-shape methods use learned prediction~\citep{wang2019shape2motion,yan2020rpm,goyal2025geopard,li2026particulate}, vision-language reasoning~\citep{qiu2025articulate}, or geometric and physical optimization~\citep{xu2026motionanymesh}.

Despite these advances, reconstructing a tabletop scene with both executable part-level articulation and single-view consistency remains challenging. Generated-mesh artifacts can corrupt joint estimation and motion-range inference, while dense arrangements and occlusions hinder input-view-consistent layout recovery. ArticuTable addresses these challenges through GRAM for executable articulated asset generation and PSGSR for progressive scene registration and pose disambiguation.

\section{Method}
\begin{figure*}[t]
    \centering
    \includegraphics[width=\textwidth]{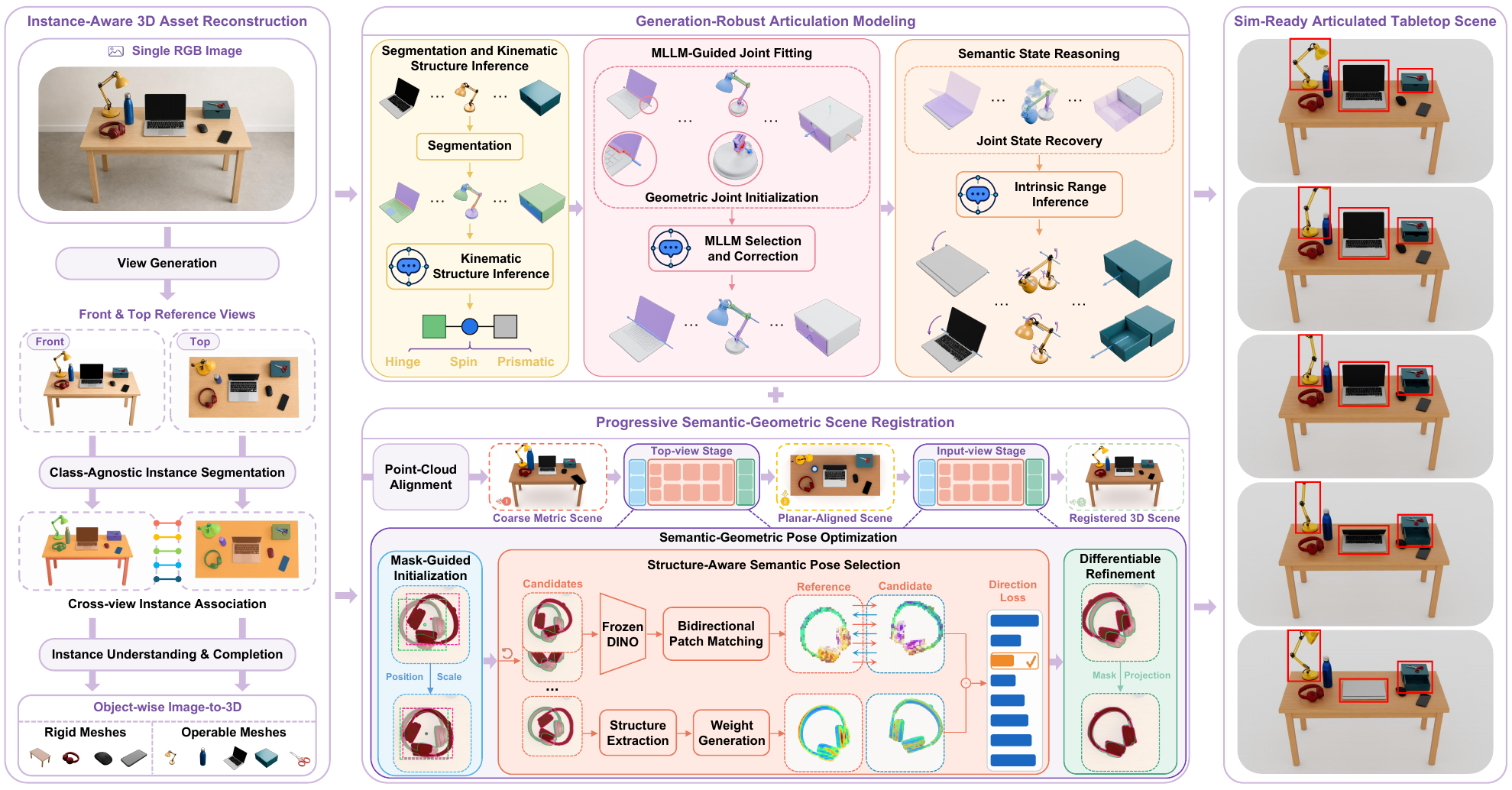}
    \caption{\textbf{Overview of ArticuTable.} From a single RGB tabletop image, ArticuTable reconstructs object-wise 3D proxies, converts operable proxies into executable URDF assets with GRAM, and registers them into a simulation-ready scene with PSGSR.}
    \label{fig:pipeline}
\end{figure*}

\subsection{Problem Setup and Method Overview}
\label{sec:overview}

Given a monocular RGB observation \(I\in\mathbb{R}^{H\times W\times 3}\), our objective is to recover a simulation-ready 3D tabletop scene that is consistent with the input view in both instance arrangement and projected appearance. Operable objects must additionally support executable part-level articulation.

Let \(G_0\) denote the supporting table and \(\{G_i\}_{i=1}^{N}\) the reconstructed object assets. We represent each asset as \(G_i=(\mathcal{L}_i,\mathcal{J}_i)\), where \(\mathcal{L}_i\) denotes its set of rigid links and \(\mathcal{J}_i\) its set of kinematic joints. Each joint encodes its parent--child connectivity, type, axis, frame, and motion limits. For notational uniformity, we treat a rigid asset as a single link with an empty joint set, whereas an operable asset comprises multiple links connected by one or more executable joints.

The reconstructed scene is represented as
\begin{equation}
    \widehat{\mathcal{S}}
    =
    \left\{
        \left(G_i,\mathbf{T}_i\right)
    \right\}_{i=0}^{N},
    \label{eq:scene_representation}
\end{equation}
where \(\mathbf{T}_i\) denotes the scene-level placement of \(G_i\). It is parameterized by a rotation \(\mathbf{R}_i\in\mathrm{SO}(3)\), an anisotropic scale matrix \(\mathbf{D}_i=\operatorname{diag}(s_i^x,s_i^y,s_i^z)\), and a translation \(\mathbf{t}_i\in\mathbb{R}^{3}\).

Figure~\ref{fig:pipeline} provides an overview of ArticuTable. ArticuTable applies instance-aware 3D asset reconstruction to obtain a gravity-aligned proxy mesh for each instance. For operable objects, GRAM recovers kinematic structures and uses MLLM-guided joint fitting and semantic state reasoning to obtain reliable joint parameters and valid motion ranges, producing executable URDF assets. For scene layout, PSGSR uses PMCR to propagate estimates across point-cloud, top-view, and input-view stages, and SASPS to resolve yaw ambiguity before differentiable refinement. The optimized placements of operable proxies are transferred to their URDF counterparts for scene assembly.

\subsection{Instance-Aware 3D Asset Reconstruction}
\label{sec:asset_reconstruction}

Semantic-first pipelines may omit objects because only previously recognized categories are passed to instance segmentation. We therefore decouple instance discovery from semantic recognition. Given the input image and a synthesized top-view hypothesis, a class-agnostic segmenter independently extracts anonymous instance proposals in both views, whose identifiers carry no category information. Front-view proposals are associated with top-view proposals through one-to-one geometric matching based on their table-normalized horizontal extents, relative spatial positions, and mask areas. An MLLM then identifies reconstructable scene instances, assigns each one a semantic identity and operability type, and estimates object dimensions, placements, and direct support relations from the paired scene observations. For each identified instance, we generate a reconstruction-ready condition image that preserves its appearance, overall pose, and articulation state, while using the matched top view only as auxiliary evidence for occluded geometry. An image-conditioned 3D generator then reconstructs an independent proxy mesh, which is standardized to the gravity-aligned coordinate frame used by GRAM and PSGSR.

\subsection{Generation-Robust Articulation Modeling}
\label{sec:gram}

For an operable object $i$, GRAM takes its monolithic proxy mesh $\mathcal{M}_i^{\mathrm{proxy}}$ as input and recovers the articulated asset $G_i=(\mathcal{L}_i,\mathcal{J}_i)$. GRAM first partitions the mesh into non-overlapping geometric primitives $\{P_k\}_{k=1}^{K}$ using 3D primitive segmentation. An MLLM then combines appearance and primitive-structure evidence to group primitives that should move rigidly together into functional links $\mathcal{L}_i$ and infer their parent--child relations and joint types, producing a rooted kinematic tree. We omit the object index $i$ from primitive-, link-, and joint-level notation when unambiguous.

\paragraph{MLLM-Guided Joint Fitting.}
Local contact geometry provides useful joint initialization, but estimates derived from distorted contact regions in image-generated meshes can be unreliable. For each predicted joint $j\in\mathcal{J}_i$ connecting a parent--child link pair $(L_p,L_c)$, GRAM approximates their contact region as
\begin{equation}
\mathcal{C}_{p,c}
=
\left\{
\mathbf{x}\in\mathcal{V}(L_c)
\mid
\min_{\mathbf{y}\in\mathcal{V}(L_p)}
\left\|\mathbf{x}-\mathbf{y}\right\|_2
\leq \delta
\right\},
\label{eq:gram_contact_set}
\end{equation}
where $\mathcal{V}(L)$ denotes the surface vertices used for contact estimation and $\delta$ is a scale-adaptive contact threshold. Geometry then provides joint-type-specific hypotheses: the contact distribution yields axis and pivot hypotheses for revolute joints, while the principal directions of the child-link geometry form axis candidates for prismatic joints. Rather than directly accepting these local estimates, the MLLM evaluates them using global multimodal and structural evidence. It selects functionally consistent motion directions and, for revolute joints, corrects the axis direction and location when supported by visible mechanical evidence. 

\paragraph{Semantic State Reasoning.}
Since a proxy mesh provides only a single observed configuration, GRAM models the observed joint state and the object-intrinsic motion range separately. For each joint, GRAM expresses its state in a canonical coordinate system and estimates the observed coordinate $q_j^{\mathrm{obs}}$ relative to the canonical zero configuration. For revolute joints, geometric closure reasoning recovers the canonical zero and observed state; for prismatic joints, the observed displacement is estimated along the fitted outward axis. Given this normalized state, semantic state reasoning combines functional semantics with geometric feasibility evidence to infer the object-intrinsic interval
\begin{equation}
\mathcal{Q}_j^{\mathrm{int}}
=
[0,q_j^{\max}],
\qquad
q_j^{\mathrm{obs}}\in\mathcal{Q}_j^{\mathrm{int}},
\label{eq:gram_intrinsic_range}
\end{equation}
where $q_j^{\max}$ denotes the maximum angular extent of a revolute joint or the maximum travel of a prismatic joint. By treating geometry as feasibility evidence rather than the sole basis for range estimation, this design reduces sensitivity to artifacts in imperfect meshes. Using the recovered canonical configuration, GRAM exports the kinematic structure, joint parameters, and intrinsic motion ranges as an executable URDF asset.

\subsection{Progressive Semantic-Geometric Scene Registration}
\label{sec:psgsr}

Given independently reconstructed proxy meshes $\{\mathcal{M}_i^{\mathrm{proxy}}\}_{i=1}^{N}$, PSGSR estimates their scene-level placements $\{\mathbf{T}_i\}_{i=1}^{N}$ in a common scene coordinate frame. For each reference view $v\in\{\mathrm{top},\mathrm{input}\}$, $I^v$, $M_i^v$, and $\boldsymbol{\Pi}_v$ denote the RGB image, instance mask, and camera, respectively. PSGSR combines PMCR with SASPS to reduce the sensitivity of scene registration to initialization.

\paragraph{Progressive Multi-Constraint Registration.}
The supporting table first establishes a shared metric scene frame. PMCR then decomposes object registration into three successive stages. Point-cloud alignment provides a coarse object placement using a trimmed nearest-surface distance. Top-view registration then refines the planar position, footprint scale, and yaw under the synthesized top-view observation, reducing the scale--depth ambiguity of perspective alignment. The support hierarchy inferred during instance understanding determines the initial vertical placement of each object. The resulting estimate initializes input-view registration, which corrects residual discrepancies in projected position, anisotropic scale, yaw, and silhouette while further refining vertical placement for object-to-object support relations. By propagating each intermediate estimate to the subsequent stage, PMCR progressively introduces metric, planar, and input-view constraints, narrowing the placement search space before each refinement.

\paragraph{Structure-Aware Semantic Pose Selection.}
Mask-guided initialization estimates position and scale without resolving yaw ambiguity. SASPS ranks yaw candidates \(\Theta_i=\{\theta_{i,k}\}\) in both stages: top-view candidates use 2D rotations, while input-view candidates are 3D renderings of residual symmetric poses. A frozen DINOv2 encoder~\citep{oquab2023dinov2} extracts patch features from the reference and rendered RGB images, from which we obtain forward and backward nearest-neighbor correspondence sets
\(\mathcal{C}_k^{\rightarrow}\) and \(\mathcal{C}_k^{\leftarrow}\).

Let \(a\) and \(b\) index reference and candidate patches, respectively. We assign their correspondence a structure-aware weight
\begin{equation}
    w_{ab}^{k}
    =
    \alpha_a^{\mathrm{ref}}
    \beta_b^{k}
    s_{ab}^{k},
    \label{eq:structure_weight}
\end{equation}
where \(\alpha_a^{\mathrm{ref}}\) measures local appearance structure in the reference image, \(\beta_b^k\) measures local geometric structure in the rendered normal map, and \(s_{ab}^k\) denotes DINO semantic matching confidence.

Since the reference and candidate crops share the same image coordinate frame, we score each candidate by the weighted displacement between matched patches:
\begin{equation}
    \mathcal{L}_{\mathrm{dir}}(\theta_{i,k})
    =
    \frac{1}{2}
    \sum_{\mathcal{C}\in
    \{\mathcal{C}_k^{\rightarrow},
      \mathcal{C}_k^{\leftarrow}\}}
    \frac{
        \sum_{(a,b)\in\mathcal{C}}
        w_{ab}^{k}
        \|\mathbf{u}_a-\mathbf{v}_b\|_2
    }{
        \sum_{(a,b)\in\mathcal{C}}w_{ab}^{k}
    },
    \label{eq:direction_loss}
\end{equation}
where \(\mathbf{u}_a\) and \(\mathbf{v}_b\) are normalized patch coordinates. The candidate with the lowest direction loss, $\theta_i^\star=\arg\min_{\theta_{i,k}\in\Theta_i}\mathcal{L}_{\mathrm{dir}}(\theta_{i,k})$, is selected as the yaw initialization.

Together with the mask-based position and scale estimates, the selected yaw hypothesis initializes differentiable refinement under a mask-projection objective that combines IoU with a boundary-aware SDF penalty.

\subsection{Construction of ArticuTable-100}
\label{sec:dataset}

We construct \textbf{ArticuTable-100}, a manually curated collection of 100 simulation-ready tabletop scenes reconstructed by ArticuTable. Candidate reconstructions are evaluated for visual fidelity, physical validity, and articulation executability. The resulting collection spans diverse tabletop arrangements, object combinations, and articulated object types. Each scene is distributed as a self-contained USDZ package directly loadable in Isaac Sim, containing textured visual assets, collision geometry for simulation, and executable articulated object models.

\section{Experiments}

\subsection{Experimental Setup}
\label{sec:experimental_setup}

We evaluate scene reconstruction and registration on 150 generated tabletop reference images against ACDC, Gen3DSR, MIDI, and TabletopGen~\citep{Dai_2025_Auto,ardelean2025gen3dsrgeneralizable3dscene,huang2025midi,wang2026tabletopgen}. All baselines share inputs and views; where applicable, they use the same TRELLIS.2~\citep{Xiang_2026_Native} and GPT-5~\citep{singh2026openaigpt5card} as ArticuTable. GRAM is evaluated on all 243 Lightwheel assets~\citep{li2026particulate} following the official protocol, covering part decomposition, articulated geometry, and joint estimation. ArticuTable uses fixed pretrained models without task-specific training or fine-tuning. Complete baseline settings, metric definitions, implementation details, and ablation protocols are provided in Appendix~\ref{app:experimental_details}.

\subsection{Scene-Level Reconstruction and Registration}
\label{sec:scene_comparison}

\begin{figure*}[t]
\centering
\includegraphics[width=\textwidth]{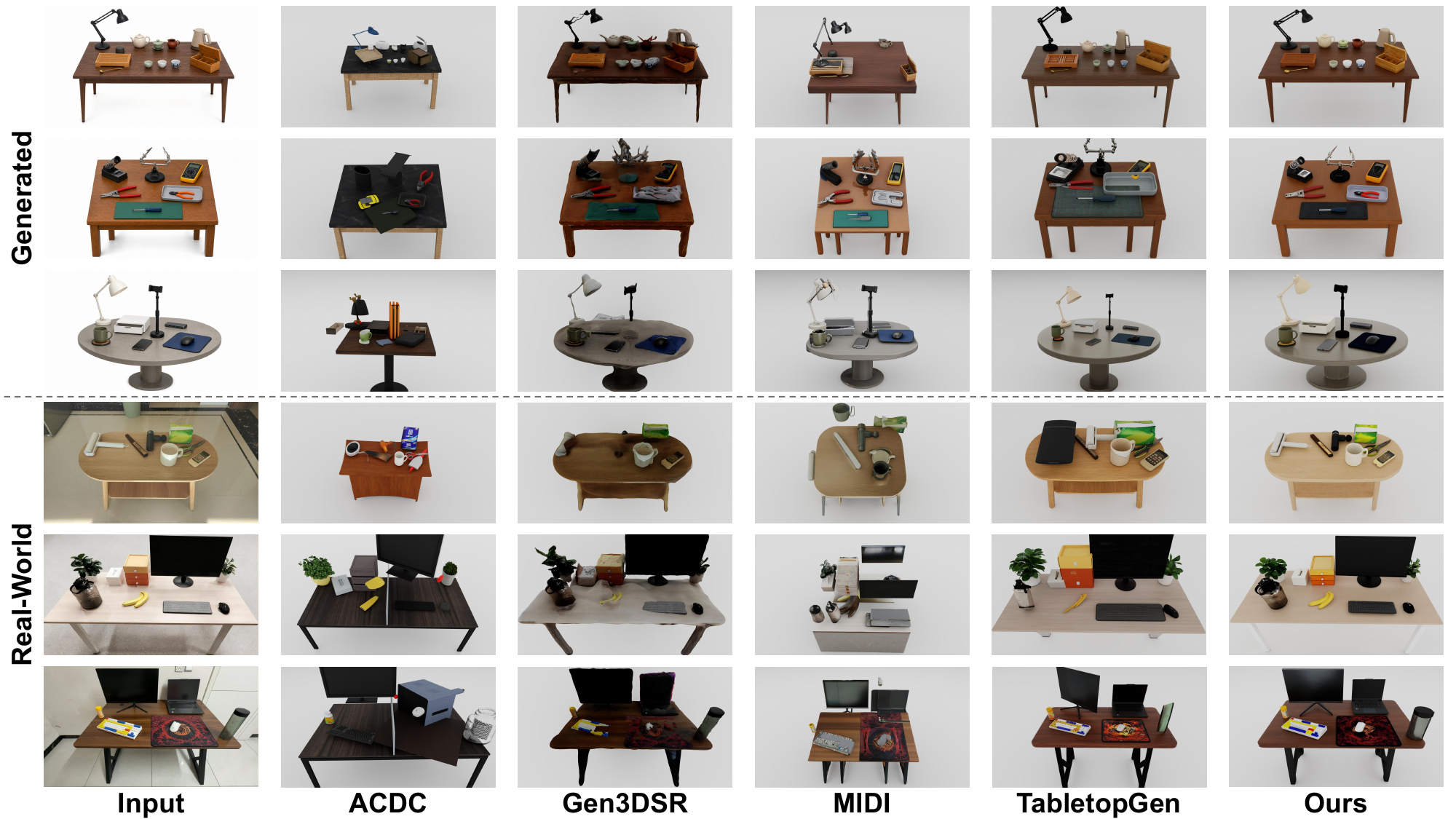}
\caption{\textbf{Qualitative comparison of scene-level reconstruction and registration on generated and real-world inputs.}}
\label{fig:scene_qualitative_comparison}
\end{figure*}

Table~\ref{tab:scene_registration_comparison} shows that ArticuTable leads across all scene-level metrics. Its advantage across image-based similarity, GPT-5 evaluation, and collision measures indicates that closer agreement with the input view is achieved without sacrificing physical validity. In particular, the substantially lower collision rates show that the recovered layouts are both visually aligned and physically plausible.

\begin{table*}[ht]
\centering
\caption{\textbf{Scene-level results.} Avg.\ is the mean of VF, IA, and PP, and OR is the mean overall rank. Best and second-best values are bolded and underlined.}
\label{tab:scene_registration_comparison}
\small
\setlength{\tabcolsep}{3pt}
\renewcommand{\arraystretch}{1.12}
\resizebox{\textwidth}{!}{%
\begin{tabular}{@{}lcccccccccc@{}}
\toprule
& \multicolumn{3}{c}{Input-View Consistency}
& \multicolumn{5}{c}{GPT-5 Evaluation}
& \multicolumn{2}{c}{Physical Validity} \\
\cmidrule(lr){2-4}
\cmidrule(lr){5-9}
\cmidrule(l){10-11}
Method
& LPIPS $\downarrow$
& DINOv2 $\uparrow$
& CLIP $\uparrow$
& VF $\uparrow$
& IA $\uparrow$
& PP $\uparrow$
& Avg. $\uparrow$
& OR $\downarrow$
& $\mathrm{Col}_{O}$ (\%) $\downarrow$
& $\mathrm{Col}_{S}$ (\%) $\downarrow$ \\
\midrule
ACDC~\citep{Dai_2025_Auto}
& 0.4447 & 0.5135 & 0.8058
& 3.927 & 1.920 & 3.880 & 3.242 & 4.298
& 1.5302 & 32.00 \\
Gen3DSR~\citep{ardelean2025gen3dsrgeneralizable3dscene}
& \underline{0.2716} & 0.5704 & 0.8386
& 2.220 & 4.680 & 4.013 & 3.638 & 3.349
& 11.3322 & 66.67 \\
MIDI~\citep{huang2025midi}
& 0.4302 & 0.6469 & 0.8668
& 2.787 & 2.933 & 2.360 & 2.693 & 4.142
& 37.9430 & 99.33 \\
TabletopGen~\citep{wang2026tabletopgen}
& 0.3935 & \underline{0.7788} & \underline{0.8908}
& \underline{5.380} & \underline{4.860} & \underline{5.893}
& \underline{5.378} & \underline{2.162}
& \underline{1.4488} & \underline{16.00} \\
\midrule
\textbf{ArticuTable (Ours)}
& \textbf{0.2398}
& \textbf{0.8991}
& \textbf{0.9265}
& \textbf{5.867}
& \textbf{6.253}
& \textbf{6.600}
& \textbf{6.240}
& \textbf{1.049}
& \textbf{0.3819}
& \textbf{10.67} \\
\bottomrule
\end{tabular}%
}
\end{table*}

Figure~\ref{fig:scene_qualitative_comparison} provides qualitative comparisons on both generated and real-world inputs. Competing methods frequently exhibit missing or distorted objects, inaccurate orientations, and noticeable deviations from the input layout. In contrast, ArticuTable reconstructs more complete tabletop scenes and more faithfully preserves object poses and spatial relationships. These results further illustrate the performance of ArticuTable across generated and real-world scenes.

\subsection{Articulated-Asset Evaluation}
\label{sec:articulation_evaluation}

Table~\ref{tab:lightwheel} evaluates GRAM's core articulation capability under the standardized Lightwheel protocol. GRAM achieves the strongest part-matching precision and joint-estimation accuracy among the evaluated mesh-conditioned methods while remaining competitive on articulated-geometry metrics. Among task-training-free approaches, its consistent gains over Articulate AnyMesh show that the improvements extend beyond joint parameters to part recovery and articulated geometry.

\begin{table}[ht]
\centering
\caption{\textbf{Results on the Lightwheel articulation benchmark~\citep{li2026particulate}.} Best and second-best results within each input modality are bolded and underlined, respectively.}
\label{tab:lightwheel}
\small
\setlength{\tabcolsep}{3.5pt}
\renewcommand{\arraystretch}{1.12}
\resizebox{\linewidth}{!}{%
\begin{tabular}{@{}llcccccccc@{}}
\toprule
Method & Input & Task-Training-Free & Prec.$\uparrow$ & Rec.$\uparrow$ & Rest mIoU$\uparrow$ & Art.\ gIoU$\uparrow$ & OC$\downarrow$ & AE ($^\circ$)$\downarrow$ & LE$\downarrow$ \\
\midrule
SINGAPO~\citep{liu2025singapo} & Image & -- & \underline{37.60} & \textbf{25.40} & \underline{0.182} & \underline{-0.263} & \textbf{0.030} & \textbf{30.30} & \textbf{0.077} \\
PAct~\citep{liu2026pact} & Image & -- & 24.60 & 16.80 & 0.092 & -0.434 & 0.056 & 42.90 & 0.188 \\
PhysX-Anything~\citep{cao2025physxanythingsimulationreadyphysical3d} & Image & -- & 20.30 & 19.20 & 0.093 & -0.459 & 0.064 & \underline{38.80} & \underline{0.123} \\
URDF-Anything+~\citep{wu2026urdfanythingautoregressivearticulated3d} & Image & -- & \textbf{70.70} & \underline{23.90} & \textbf{0.260} & \textbf{-0.138} & \underline{0.033} & 50.40 & 0.128 \\
\midrule
Articulate AnyMesh~\citep{qiu2025articulate} & Mesh & $\checkmark$ & 89.20 & 42.50 & 0.452 & 0.158 & \underline{0.010} & 21.70 & 0.043 \\
URDF-Anything+~\citep{wu2026urdfanythingautoregressivearticulated3d} & Mesh & -- & 77.50 & 27.40 & 0.267 & -0.110 & 0.026 & 49.90 & 0.111 \\
Particulate~\citep{li2026particulate} & Mesh & -- & \underline{89.90} & \textbf{51.50} & \textbf{0.576} & \textbf{0.305} & \textbf{0.009} & \underline{20.90} & \underline{0.040} \\
\midrule
\textbf{GRAM (Ours)} & Mesh & $\checkmark$ & \textbf{91.15} & \underline{46.54} & \underline{0.564} & \underline{0.283} & 0.014 & \textbf{17.26} & \textbf{0.034} \\
\bottomrule
\end{tabular}%
}
\vspace{2pt}
\footnotesize
\textit{Task-Training-Free} indicates that no articulation-specific training or fine-tuning is required.
\end{table}

\begin{figure*}[ht]
\centering
\includegraphics[width=\textwidth,trim=3pt 3pt 3pt 3pt,clip]{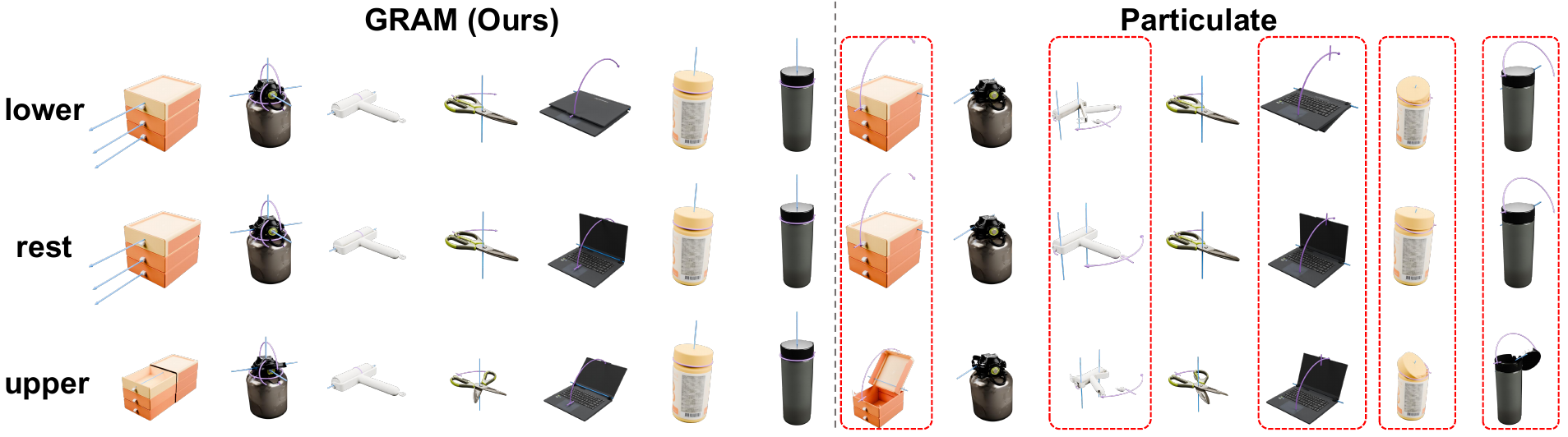}
\caption{\textbf{Robustness on raw image-to-3D meshes.} Particulate versus GRAM across motion states.}
\label{fig:particulate_robustness}
\end{figure*}

Although evaluation on Lightwheel quantifies articulation accuracy under standardized benchmark conditions, it does not reflect the mesh imperfections encountered in our image-to-3D pipeline. To assess performance under these practical conditions, we manually audit all 425 candidate assets across the complete 150-scene evaluation set. Of these, 418 are judgeable, with 346 (82.8\%) passing both structural and motion criteria; the remaining seven lack sufficient visual evidence and are treated as unjudgeable (Appendix~\ref{app:generated_articulation_audit}; Table~\ref{tab:generated_articulation_audit}). We further compare GRAM with Particulate~\citep{li2026particulate}, a strong Lightwheel baseline, on every articulated object reconstructed from the presented real scenes. As shown in Fig.~\ref{fig:particulate_robustness}, GRAM maintains coherent geometry across motion states, whereas Particulate can produce implausible motion or degenerate states. Figures~\ref{fig:teaser} and~\ref{fig:additional_articulated_scenes} further illustrate part-level articulation in reconstructed tabletop scenes.

\subsection{Ablation Studies}
\label{sec:ablation_studies}

\begin{table}[ht]
\centering
\small
\renewcommand{\arraystretch}{1.12}
\caption{\textbf{GRAM component ablation on Lightwheel.} Enabled and disabled values are averaged over both settings of the other component; lower is better.}
\label{tab:component_ablation}
\begin{tabular*}{\linewidth}{@{\extracolsep{\fill}}llccc@{}}
\toprule
Component & Metric & Enabled & Disabled & Reduction \\
\midrule
\multirow{2}{*}{MLLM-Guided Joint Fitting}
& AE $\downarrow$ & $\mathbf{17.17}^{\circ}$ & $40.06^\circ$ & 57.1\% \\
& LE $\downarrow$ & \textbf{0.0417} & 0.0527 & 20.9\% \\
\midrule
\multirow{2}{*}{Semantic State Reasoning}
& Art.\ PC $\downarrow$ & \textbf{0.2502} & 0.2605 & 3.9\% \\
& OC $\downarrow$ & \textbf{0.0152} & 0.0176 & 13.6\% \\
\bottomrule
\end{tabular*}
\end{table}

Table~\ref{tab:component_ablation} confirms the complementary roles of the two GRAM components. MLLM-guided joint fitting primarily improves joint-axis direction and location, supporting the use of global multimodal evidence to correct unreliable local estimates. Semantic state reasoning improves articulated geometry across motion states, showing that semantic range estimation helps avoid implausible configurations caused by unreliable mesh-derived limits.

\begin{table*}[ht]
\centering
\caption{\textbf{PSGSR ablation.} ``--'' denotes a component not applicable to the corresponding variant.}
\label{tab:registration_ablation}
\small
\renewcommand{\arraystretch}{1.12}
\resizebox{\textwidth}{!}{%
\begin{tabular}{@{}lcccccc@{}}
\toprule
Variant
& Top-view Stage
& Input-view Stage
& SASPS
& LPIPS $\downarrow$
& DINOv2 $\uparrow$
& CLIP $\uparrow$ \\
\midrule
Point-cloud alignment only
& -- & -- & --
& 0.3439 & 0.8102 & 0.9077 \\
w/o Top-view stage
& $\times$ & $\checkmark$ & --
& 0.2564 & 0.8702 & \underline{0.9210} \\
Top-view stage only
& $\checkmark$ & $\times$ & $\times$
& 0.3235 & 0.8586 & 0.9156 \\
w/o SASPS
& $\checkmark$ & $\checkmark$ & $\times$
& \underline{0.2543} & \underline{0.8762} & 0.9205 \\
Full model
& $\checkmark$ & $\checkmark$ & $\checkmark$
& \textbf{0.2398} & \textbf{0.8991} & \textbf{0.9265} \\
\bottomrule
\end{tabular}%
}
\end{table*}

Table~\ref{tab:registration_ablation} shows that view-based refinement consistently improves upon coarse point-cloud alignment. With SASPS disabled, adding the top-view stage improves LPIPS and DINOv2 while maintaining comparable CLIP similarity, supporting the progressive top-view-to-input-view design of PMCR. Enabling SASPS yields the best results across all three metrics, confirming that structure-aware semantic pose selection provides a stronger yaw initialization for differentiable refinement.

\begin{figure*}[ht]
    \centering
    \includegraphics[width=\textwidth]{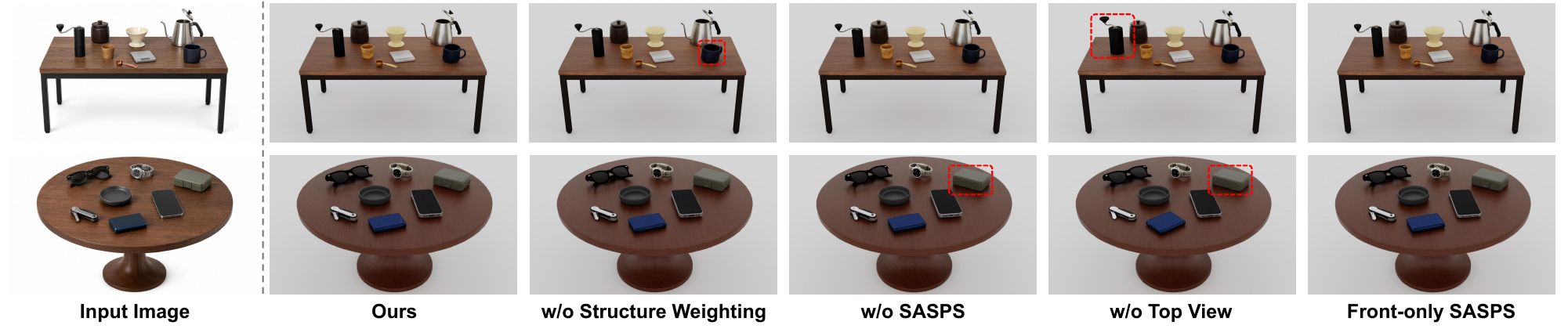}
    \caption{Qualitative pose-selection ablation. Red boxes indicate representative orientation errors.}
    \label{fig:qualitative_pose_selection_ablation}
\end{figure*}

Scene-level metrics such as LPIPS, DINOv2, and CLIP primarily measure overall image similarity, so local orientation discrepancies may have limited effects on their scores, and an incorrect orientation can sometimes score higher. To directly examine these local differences, Fig.~\ref{fig:qualitative_pose_selection_ablation} provides qualitative comparisons under different component ablations: removing structure weighting selects a handle direction inconsistent with the reference, while removing SASPS reverses the orientation of the latched case. Front-only SASPS achieves orientation results comparable to the top-view-assisted setting, but incurs substantially higher computational cost because every hypothesis requires a new 3D rendering. The average optimization time more than doubles from 179.4 to 374.8 seconds ($2.09\times$), reaching 595.3 seconds ($3.32\times$) with sequential rendering. In contrast, the top-view stage enables efficient orientation search through inexpensive 2D rotations.

\subsection{User Study}
\label{sec:user_study}

\begin{table}[ht]
\centering
\caption{\textbf{User-study rankings from 59 participants.} Each method receives 590 valid rankings; best and second-best results are bolded and underlined.}
\label{tab:user_study}
\small
\renewcommand{\arraystretch}{1.05}
\begin{tabular*}{\linewidth}{@{\extracolsep{\fill}}lccccc@{}}
\toprule
Metric
& ACDC
& Gen3DSR
& MIDI
& TabletopGen
& \textbf{ArticuTable} \\
\midrule
Mean Rank $\downarrow$
& 4.08 & 3.75 & 3.75 & \underline{2.24} & \textbf{1.17} \\
Median Rank $\downarrow$
& 4 & 4 & 4 & \underline{2} & \textbf{1} \\
First-Place Rate $\uparrow$
& 0.17\% & 1.02\% & 1.02\% & \underline{12.20\%} & \textbf{85.59\%} \\
Top-2 Rate $\uparrow$
& 6.61\% & 11.86\% & 8.81\% & \underline{74.58\%} & \textbf{98.14\%} \\
\bottomrule
\end{tabular*}
\end{table}

Table~\ref{tab:user_study} shows a strong and consistent preference for ArticuTable. It is ranked first in 85.59\% of the evaluations and leads all summary measures by a wide margin over the strongest baseline, TabletopGen. Its near-universal top-two placement indicates that this preference is consistent rather than driven by a small subset of evaluations.

\subsection{Simulation-Ready Validation}
\label{sec:simulation_validation}

\begin{figure*}[t]
\centering
\includegraphics[width=\textwidth]{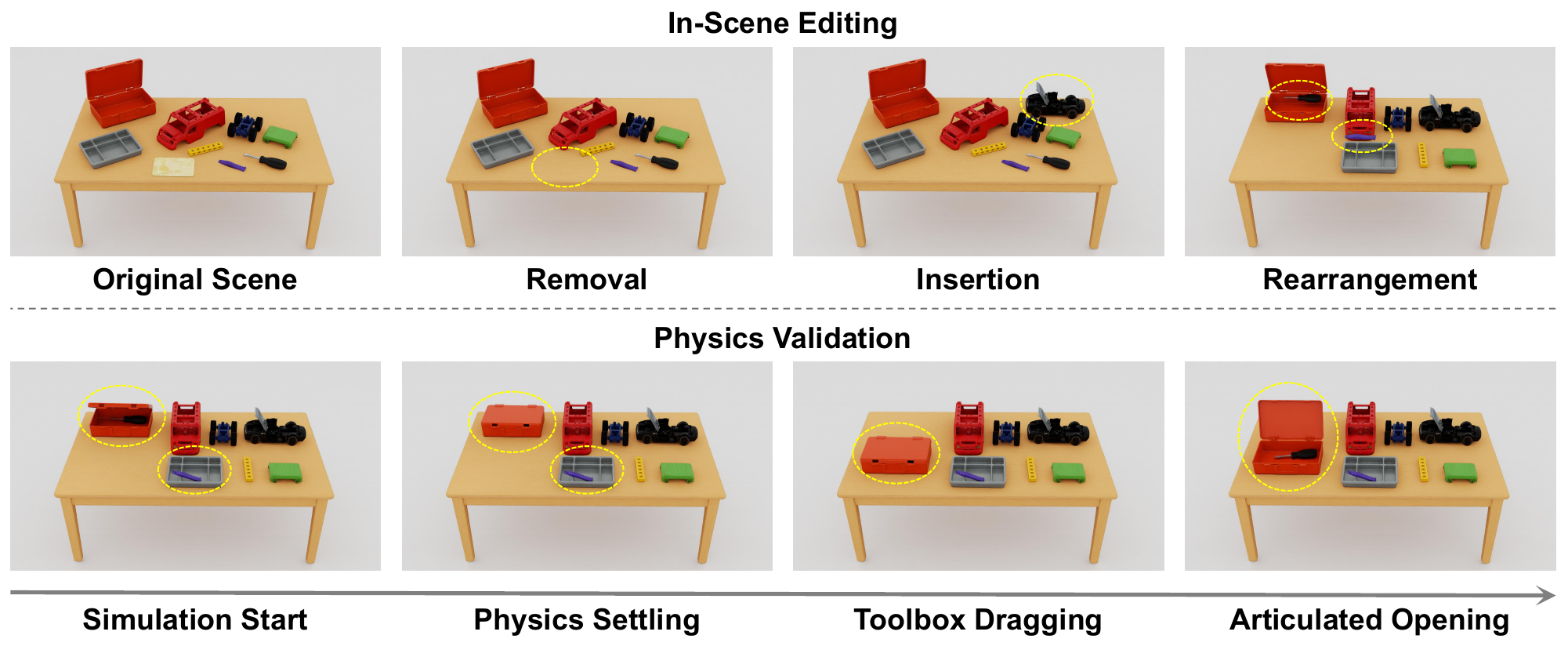}
\caption{\textbf{Scene editing and physical interaction in Isaac Sim.} Top: object removal, insertion, and rearrangement. Bottom: gravity settling, contact-based transport, and articulated opening. Yellow dashed ellipses highlight the edited or manipulated regions.}
\label{fig:simready_validation}
\end{figure*}

Figure~\ref{fig:simready_validation} demonstrates scene editing and physics-based interaction in Isaac Sim. The instance-level representation supports object removal, insertion, and rearrangement. In the representative interaction sequence, objects settle under gravity and contact; moving the toolbox transports its unattached contents through collision; and the recovered lid joint remains operable after relocation. These results demonstrate usable collision geometry and executable articulation in the assembled scene. Export settings and validation procedures are detailed in Appendix~\ref{app:simready_validation}.

\section{Limitation}
\begin{figure}[ht]
\centering
\includegraphics[width=\linewidth]{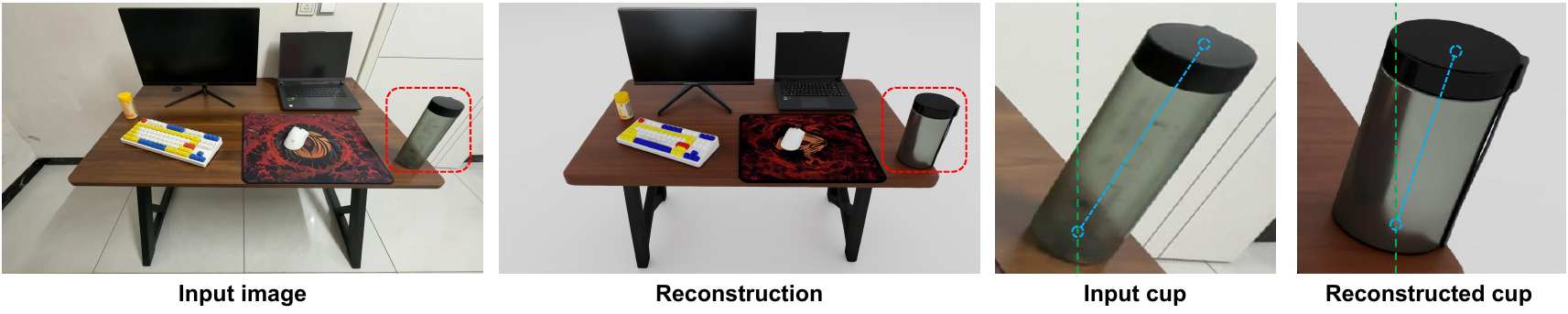}
\caption{\textbf{Residual projection misalignment.} Despite the overall scene alignment, the reconstructed cup exhibits less projected tilt than the input cup. Red boxes identify the cups; cyan dashed lines indicate their projected axes, and green dashed lines indicate the image vertical.}
\label{fig:limitation_cup}
\end{figure}

ArticuTable renders the reconstructed scene with an estimated pinhole camera, whose projection model and parameters may not fully match those of the input image. Under strong off-axis viewing, such discrepancies can become more visible near image boundaries. As shown in Fig.~\ref{fig:limitation_cup}, the reconstructed cup exhibits less projected tilt than the input cup despite the overall scene alignment. Future work could jointly refine camera parameters and object transformations while explicitly accounting for lens distortion.

\section{Conclusion}

We presented \textbf{ArticuTable}, a framework for reconstructing simulation-ready tabletop scenes containing both rigid and articulated objects from a single RGB image. GRAM transforms monolithic object meshes into executable articulated assets by recovering functional links, rooted kinematic structures, and joint parameters. PSGSR improves scene registration through progressively constrained metric, planar, and input-view stages, while SASPS selects reliable yaw hypotheses for differentiable refinement. Experiments demonstrate improvements in scene-level visual consistency, physical validity, articulation quality, and joint estimation. We further introduced \textbf{ArticuTable-100}, a curated collection of 100 simulation-ready scenes for downstream embodied manipulation. Together, these contributions provide a unified basis for constructing visually faithful and physically interactive tabletop environments from monocular observations.

\bibliography{iclr2027_conference}
\bibliographystyle{iclr2027_conference}

\newpage

\appendix

\section{Additional Details of Instance-Aware 3D Asset Reconstruction}
\label{app:asset_reconstruction}

\subsection{Anonymous Instance Discovery}

Given the input image and its synthesized top-view hypothesis, we independently extract anonymous instance proposals from the two views. Instead of querying a predefined category list, we use the single generic concept ``tabletop item,'' while detecting the complete table and tabletop surface separately as spatial supports. The resulting identifiers represent only view-local mask proposals and contain no semantic category information. The scene inventory remains input-view-driven: every retained input-view proposal is passed to subsequent instance understanding, while top-view proposals provide auxiliary cross-view evidence.

\subsection{Geometric Cross-View Association}

We associate input-view and top-view proposals before assigning semantic identities. Let $A_i^v$ and $B_i^v=[x_i^{l,v},y_i^{t,v},x_i^{r,v},y_i^{b,v}]$ denote the mask and bounding box of proposal $i$ in view $v\in\{\mathrm{input},\mathrm{top}\}$. Using the detected table ($i=0$) as a common reference, we define the table-normalized horizontal extent
\begin{equation}
\mathbf{r}_i^v=\left(\frac{x_i^{l,v}-x_0^{l,v}}{w_0^v},\frac{x_i^{r,v}-x_0^{l,v}}{w_0^v}\right),\qquad w_0^v=x_0^{r,v}-x_0^{l,v},
\label{eq:anonymous_horizontal_extent}
\end{equation}
and the relative mask area
\begin{equation}
a_i^v=\frac{|A_i^v|}{|A_0^v|}.
\label{eq:anonymous_relative_area}
\end{equation}
We additionally define $d_i^v$ by min--max normalizing the bottom coordinate $y_i^{b,v}$ over all non-table proposals in the same view, providing a view-relative spatial-position descriptor.

For an input-view proposal $i$ and a top-view proposal $j$, the association cost is
\begin{equation}
c_{ij}=\left\|\mathbf{r}_i^{\mathrm{input}}-\mathbf{r}_j^{\mathrm{top}}\right\|_1+\lambda_d\left|d_i^{\mathrm{input}}-d_j^{\mathrm{top}}\right|+\left|\sqrt{a_i^{\mathrm{input}}}-\sqrt{a_j^{\mathrm{top}}}\right|.
\label{eq:anonymous_cross_view_cost}
\end{equation}
We set $\lambda_d=0.25$. As a sensitivity check, removing the relative-position term by setting $\lambda_d=0$ changes 28 of the 1,184 associations across 150 scenes, with $97.64\%$ remaining unchanged. Increasing $\lambda_d$ to $1$ changes only 2 associations, with $99.83\%$ remaining unchanged. These results indicate that the matching is robust to the precise value of $\lambda_d$, while the relative-position term affects a small subset of ambiguous associations.

We obtain a one-to-one association by minimizing the total cost with the Hungarian algorithm~\citep{munkres1957algorithms}. Because the matching operates only on table-normalized geometry and mask statistics, it does not require category agreement between views. Unmatched input-view proposals remain in the scene inventory and are processed using input-view evidence alone, whereas unmatched top-view proposals are not introduced as additional scene instances.

\subsection{Instance Understanding and Object-Wise Reconstruction}

Semantic reasoning is performed after anonymous proposals have been discovered and associated. An MLLM examines each input-view proposal, determines whether it represents a complete and independently reconstructable scene instance, and assigns each retained instance a semantic identity and operability type. At the scene level, the MLLM further estimates object dimensions, planar placements, orientations, and direct support relations from the input and top-view observations, producing a support hierarchy rooted at the table. When a cross-view match is available, the corresponding top-view observation provides complementary evidence for object footprint and occluded geometry without overriding the appearance or physical state observed in the input image.

For every retained instance, we construct a target-specific condition image in which the input observation provides the primary evidence for the object's appearance, overall pose, and articulation state, while the matched top view is used only to complete occluded geometry. TRELLIS.2~\citep{Xiang_2026_Native} reconstructs each condition image as an independent proxy mesh. The reconstructed meshes are then centered and gravity-aligned, while the inferred support hierarchy is passed to PSGSR for scene-level placement.

\section{Additional Details of Generation-Robust Articulation Modeling}
\label{app:gram}

\subsection{Segmentation and Kinematic Structure Inference}

GRAM receives a textured monolithic mesh generated for an operable scene instance. Let $\mathbf{c}$ be the center of its axis-aligned bounding box and $D$ its bounding-box diagonal. Following the gravity alignment described in Sec.~\ref{sec:overview}, GRAM centers and isotropically scales the mesh within a right-handed Y-up working frame by
\begin{equation}
    \mathbf{x}' = \frac{\mathbf{x}-\mathbf{c}}{D}.
    \label{eq:gram_normalization}
\end{equation}
Degenerate faces are removed, and P3-SAM~\citep{ma2025p3sam} assigns one primitive label to every remaining face. Primitive IDs are reordered by decreasing surface area, and the labels are transferred back to the textured source geometry. This produces a textured working mesh together with a dense face-to-primitive map.

GRAM next renders five orthographic views from the front, right, back, left, and top. Each camera produces an RGB rendering and a primitive-ID rendering with identical projection parameters. The RGB views expose appearance and functional cues, whereas the ID views establish an exact correspondence between visible regions and 3D primitives.

The MLLM uses the paired views, primitive mesh, source and primary images, and boundary evidence to group primitives that undergo rigid co-motion. It predicts a set of links, their parent--child relations, and a joint type for every non-root link. Revolute joints are further classified as \emph{hinge} or \emph{spin}. Each link is materialized as a textured visual mesh and a convex-hull collision mesh.

\subsection{MLLM-Guided Joint Fitting}

For each joint $j=(L_p,L_c)$, GRAM approximates the contact region using child surface vertices whose nearest parent surface lies within a scale-adaptive distance, preferring boundary vertices when available. The complete subtree rooted at $L_c$ is treated as the moving body, while all links outside that subtree form the obstacle set. Consequently, joint fitting and collision evaluation follow the predicted kinematic hierarchy rather than moving the child link in isolation.

For a hinge joint, the contact centroid initializes the pivot and the largest-variance principal direction of the contact set initializes the axis. For a spin joint, GRAM estimates the contact plane by PCA, fits a circle in that plane using seeded RANSAC~\citep{fischler1981random} followed by least squares, and lifts the circle center and plane normal back to 3D. For a prismatic joint, PCA of the child vertices yields three candidate axes. The MLLM selects one candidate and its positive direction using the paired renderings and the source image. For revolute joints, the MLLM evaluates the contact-derived axis line using the paired views, source image, link geometry, and global kinematic structure. It corrects the axis direction when supported by explicit visual or mechanical evidence and estimates the pivot location from the visible attachment and rotation structure. When sufficient directional evidence is unavailable, the contact-derived axis direction is retained.

\subsection{Semantic State Reasoning and Intrinsic Limits}

Joint geometry alone does not identify which configuration of an image-generated mesh should be treated as the semantic zero. GRAM therefore separates geometric closure recovery from semantic extent reasoning. For hinge joints, connected planar patches on the parent and child links generate candidate closed configurations around the fitted axis, while the complete moving subtree participates in geometric and collision-based validation. Candidates are ranked using plane alignment, overlap, axis proximity, and geometric consistency, yielding a plane-closed configuration and the signed displacement of the reconstructed mesh from that configuration. For spin joints, GRAM searches for the nearest closed configuration in both rotational directions while moving the complete child subtree. Axis-local surfaces within an adaptive neighborhood of the rotation axis are excluded from the collision predicate, preventing axle-like contact from blocking an otherwise valid rotation.

Prismatic joints use their selected outward direction to define a nonnegative coordinate from the canonical closed state. Motion is scanned in both directions until collision or loss of parent contact; a displacement is considered derailed when fewer than $5\%$ of the transformed contact samples remain close to the parent. These prismatic scans provide physical fallback intervals and geometric feasibility evidence.

Given the recovered canonical state, semantic state reasoning infers the object-intrinsic interval
$\mathcal{Q}_j^{\mathrm{int}}=[0,q_j^{\max}]$ by combining functional semantics with geometric feasibility evidence. For revolute joints, the MLLM uses the primary image, calibrated joint renderings, recovered canonical configuration, and global kinematic structure to infer the mechanism-specific maximum angular extent. For prismatic joints, it combines the source and primary images, paired multiview renderings, preserved reconstructed displacement, and physical scan evidence to infer the maximum travel. Throughout this process, the fitted axis, pivot, topology, and positive motion direction remain fixed.

GRAM records three state quantities for every joint:
\begin{equation}
    q_j^{\mathrm{zero}}=-q_j^{\mathrm{mesh}},
    \qquad
    q_j^{\mathrm{mesh}}\equiv q_j^{\mathrm{obs}},
    \qquad
    q_j^{\mathrm{mesh}},q_j^{\mathrm{scene}}
    \in \mathcal{Q}_j^{\mathrm{int}}=[0,q_j^{\max}],
    \label{eq:gram_state_contract}
\end{equation}
where $q_j^{\mathrm{zero}}$ is the transform baked into the link geometry, $q_j^{\mathrm{mesh}}$ is the state represented by the reconstructed proxy mesh and corresponds to $q_j^{\mathrm{obs}}$ in Sec.~\ref{sec:gram}, and $q_j^{\mathrm{scene}}$ is the state used when the asset is instantiated in the assembled scene.

\subsection{URDF Export and Scene-Conditioned Motion}

Before export, GRAM verifies a one-to-one correspondence between predicted links, fitted joints, and state metadata. It bakes the semantic zero transform into each child subtree, converts joint axes and pivots from the normalized Y-up working frame to a right-handed Z-up URDF frame, and writes textured GLB visual meshes together with OBJ collision meshes. Link mass and inertia are computed from the collision volume using a fixed material density. The URDF contains the validated parent--child tree, joint types, axes, frames, and intrinsic limits; a companion manifest stores the working-to-URDF coordinate-frame conversion and the zero, mesh, and scene state of every joint. Generated XML is parsed again and every referenced mesh path is checked before the asset is accepted.

The exported URDF limits describe object-intrinsic motion. After PSGSR places the articulated asset in the reconstructed tabletop scene, a separate scene-level collision pass computes the collision-free subintervals permitted by the surrounding objects. These scene-conditioned intervals are recorded separately, leaving the object-intrinsic URDF limits and recovered internal kinematics unchanged.

\section{Additional Details of Progressive Semantic-Geometric Scene Registration}
\label{app:psgsr}

\subsection{Table Alignment and Metric Scene Frame}
PSGSR uses the supporting table to place the input-view reconstruction, camera, synthesized top view, and reconstructed assets in a common metric coordinate frame. Given the input image, VGGT~\citep{Wang_2025_VGGT} estimates a dense point map $\mathcal{P}^{\mathrm{vggt}}$ together with a perspective camera $\boldsymbol{\Pi}_{\mathrm{input}}^{\mathrm{vggt}}$. Both are expressed in a shared VGGT coordinate frame whose metric scale is unknown. Let $\mathcal{P}_0^{\mathrm{vggt}}\subset\mathcal{P}^{\mathrm{vggt}}$ denote the points selected by the tabletop mask, and let $W_0$ denote the physical table width estimated during instance understanding.

We fit a plane to $\mathcal{P}_0^{\mathrm{vggt}}$ and construct a rotation $\mathbf{R}_0$ that aligns its normal with the world vertical axis while orienting the horizontal axes using the input-camera viewing direction projected onto the tabletop. The estimated table width then resolves the unknown metric scale:
\begin{equation}
s_0
=
\frac{W_0}
{\operatorname{extent}_x
\left(
\mathbf{R}_0\mathcal{P}_0^{\mathrm{vggt}}
\right)},
\qquad
\mathbf{p}^{\mathrm{world}}
=
s_0\mathbf{R}_0\mathbf{p}^{\mathrm{vggt}}
+
\mathbf{t}_0,
\label{eq:table_metric_transform}
\end{equation}
where $\operatorname{extent}_x(\cdot)$ denotes the robust tabletop extent along the width direction. The translation $\mathbf{t}_0$ places the tabletop center at the scene origin and its supporting surface at $z=0$. The same similarity transformation converts the VGGT object point clouds and the extrinsics of $\boldsymbol{\Pi}_{\mathrm{input}}^{\mathrm{vggt}}$ into the metric scene frame, while the input-camera intrinsics remain unchanged.

We subsequently align the metric table proxy with the transformed tabletop points to refine its horizontal orientation and translation, and further align its projected geometry with the detected table regions in the input and synthesized top views. For the synthesized top view, we define a vertical orthographic camera $\boldsymbol{\Pi}_{\mathrm{top}}$ centered according to the detected table region. If the table spans $w_0^{\mathrm{top}}$ pixels in this view, its metric pixel size is
\begin{equation}
r_{\mathrm{top}}
=
\frac{W_0}{w_0^{\mathrm{top}}},
\label{eq:top_view_metric_scale}
\end{equation}
which converts top-view mask coordinates and extents into metric planar positions and footprint scales. The calibrated cameras, table, and VGGT point clouds therefore share the same table-anchored metric scene frame used by all subsequent registration stages.

\subsection{Progressive Multi-Constraint Registration}

\paragraph{Point-Cloud Alignment.}
For object \(i\), let \(\mathcal{X}_i\) denote surface samples from its gravity-aligned proxy mesh. We obtain the observed object point cloud \(\mathcal{P}_i\) by selecting valid VGGT points within its input-view instance mask \(M_i^{\mathrm{input}}\) and transforming them into the metric scene frame using Eq.~\ref{eq:table_metric_transform}. At this stage, the placement is parameterized by planar translation \(\mathbf{t}_i^{xy}=(t_i^x,t_i^y,0)^\top\), yaw \(\theta_i\), and uniform scale \(s_i\). For each observed point \(\mathbf{p}\in\mathcal{P}_i\), we compute
\begin{equation}
d_i(\mathbf{p})
=
\min_{\mathbf{x}\in\mathcal{X}_i}
\left\|
\mathbf{p}
-
\left(
s_i\mathbf{R}_z(\theta_i)\mathbf{x}
+
\mathbf{t}_i^{xy}
\right)
\right\|_2.
\label{eq:psgsr_nearest_surface}
\end{equation}
Let \(\mathcal{I}_i\subseteq\mathcal{P}_i\) denote the inlier points retained after removing the largest nearest-surface residuals. The trimmed point-cloud objective is
\begin{equation}
\mathcal{L}_{\mathrm{pc}}^i
=
\frac{1}{|\mathcal{I}_i|}
\sum_{\mathbf{p}\in\mathcal{I}_i}
d_i(\mathbf{p}).
\label{eq:psgsr_trimmed_pointcloud}
\end{equation}
Trimming reduces the influence of depth outliers, occluded surfaces, and geometry absent from the reconstructed proxy. Minimizing this objective yields the coarse estimate \(\mathbf{T}_i^{\mathrm{pc}}\). This stage refines only planar translation, yaw, and uniform scale.

\paragraph{Top-View and Input-View Registration.}
The point-cloud estimate initializes top-view registration, which refines metric planar position, footprint scale, and yaw against the synthesized top-view observation. The resulting estimate initializes input-view registration, which further refines planar position, anisotropic scale, yaw, and silhouette consistency. The estimates are propagated as
\begin{equation}
\mathbf{T}_i^{\mathrm{pc}}
\longrightarrow
\mathbf{T}_i^{\mathrm{top}}
\longrightarrow
\mathbf{T}_i^{\mathrm{input}}.
\label{eq:psgsr_progressive_estimates}
\end{equation}
This propagation retains the table-anchored metric initialization while progressively introducing top-view planar evidence and input-view projection evidence.

\paragraph{Support-Aware Vertical Placement.}
The inferred support hierarchy is traversed from the table upward. Objects directly supported by the table remain grounded at \(z=0\). For an object supported by another object, its initial vertical position \(z_i^0\) is set to the estimated upper-surface height of its supporting parent.

During input-view refinement, PSGSR optimizes a vertical residual \(\eta_i\) together with the object's planar pose and scale. For a lower-surface point \(\mathbf{x}\) of the child object, let \(h_p(\mathbf{x}^{xy})\) denote the highest intersection between the vertical line through \(\mathbf{x}^{xy}\) and the triangular surface of a supporting parent \(p\). The contact offset required by that parent is estimated from
\begin{equation}
\Delta z_{i,p}
=
\max_{\mathbf{x}\in\mathcal{B}_{i,p}}
\left[
h_p(\mathbf{x}^{xy})-\mathbf{x}^{z}
\right],
\label{eq:support_contact_offset}
\end{equation}
where \(\mathcal{B}_{i,p}\) contains lower child-surface points whose horizontal coordinates overlap the supporting parent. The offsets from all declared parents define a common contact-feasible interval \(\mathcal{Z}_i\). The applied vertical position is
\begin{equation}
z_i
=
z_i^0
+
\operatorname{Proj}_{\mathcal{Z}_i}
\left(
\eta_i
\right),
\label{eq:support_aware_vertical_placement}
\end{equation}
where \(\operatorname{Proj}_{\mathcal{Z}_i}\) restricts the optimized residual to vertical offsets that preserve contact with the supporting surfaces. The interval is recomputed from the current child and parent geometry during refinement. If the contact requirements of multiple supports do not admit a common interval, their violation is penalized in the optimization objective. This allows input-view refinement to adjust the vertical placement of object-supported instances while discouraging floating and interpenetration.

\subsection{Structure-Aware Semantic Pose Selection}

\paragraph{Yaw Hypothesis Sampling.}
PSGSR estimates the relative dominant-axis alignment between the rendered and reference masks. When the mask axes provide reliable directional cues, yaw hypotheses are sampled around the two opposite axis alignments; otherwise, the full yaw range is searched. The resulting hypotheses are passed to SASPS for semantic ranking.

\paragraph{Structure-Weighted Semantic Correspondence.}
For each candidate in \(\Theta_i\), SASPS renders an RGB image, a foreground mask, and a surface-normal map. All candidate renderings are cropped using a square region determined by the reference mask, thereby preserving a shared image coordinate frame. The crops are then encoded by a frozen DINOv2 ViT-L/14 encoder~\citep{oquab2023dinov2} to obtain normalized patch features. DINOv2 is applied only to the reference and rendered RGB crops; rendered normal maps are used to compute geometric structure weights rather than as encoder inputs.

The reference appearance weight $\alpha_a^{\mathrm{ref}}$ and candidate geometry weight $\beta_b^k$ are computed from local variations in the reference RGB image and rendered surface-normal map, respectively. Bidirectional nearest-neighbor matching between normalized DINOv2 patch features produces the forward and backward correspondence sets used in Eq.~\ref{eq:direction_loss}.

\subsection{Mask-Projection Refinement}

Let \(\widehat{M}_i^v\in[0,1]\) denote the differentiably rendered silhouette of object \(i\) in view \(v\), and let \(M_i^v\) denote its target mask. We refine the selected placement using
\begin{equation}
\mathcal{L}_{\mathrm{proj}}^v
=
1-
\operatorname{IoU}
\left(
\widehat{M}_i^v,M_i^v
\right)
\exp
\left[
-\mathcal{L}_{\mathrm{sdf}}
\left(
\widehat{M}_i^v,M_i^v
\right)
\right],
\label{eq:psgsr_projection_loss}
\end{equation}
where \(\mathcal{L}_{\mathrm{sdf}}\) is a boundary-aware distance penalty that penalizes extraneous rendered regions and missing target regions according to their distances from the target boundary. The IoU term encourages global overlap, while the SDF term provides useful boundary guidance when the rendered and target silhouettes overlap only partially. In practice, the stage-specific data terms are supplemented with lightweight regularization penalties on pose and scale drift and support feasibility to stabilize optimization.

\section{Additional Experimental Details}
\label{app:experimental_details}

\subsection{Scene-Level Evaluation Protocol}

The scene-level evaluation set contains 150 tabletop reference images produced by an image-generation model from diverse tabletop-scene descriptions. These images are generated directly in the image domain rather than rendered from paired ground-truth 3D scenes. Each method receives exactly the same monocular reference image, and its reconstructed scene is rendered from the corresponding input viewpoint. We compare ArticuTable with ACDC~\citep{Dai_2025_Auto}, Gen3DSR~\citep{ardelean2025gen3dsrgeneralizable3dscene}, MIDI~\citep{huang2025midi}, and TabletopGen~\citep{wang2026tabletopgen}.

For a controlled comparison, we standardize shared upstream components only where compatible with each baseline's design. Baselines requiring object-wise 3D generation use the same TRELLIS.2 model~\citep{Xiang_2026_Native} as ArticuTable, whereas retrieval-based methods retain their original asset-retrieval mechanisms. Baselines requiring multimodal reasoning use the same GPT-5 model~\citep{singh2026openaigpt5card} as ArticuTable. All remaining method-specific stages follow the respective reproduced pipelines. This protocol controls shared foundation models without altering the defining generation or retrieval paradigm of each method.

Input-view consistency is measured using LPIPS~\citep{zhang2018unreasonable}, DINOv2~\citep{oquab2023dinov2} similarity, and CLIP~\citep{radford2021learning} similarity between the reference image and the reconstructed-scene rendering. Following TabletopGen, GPT-5 evaluates Visual Fidelity (VF), Image Alignment (IA), and Physical Plausibility (PP) on a 1--7 scale and assigns an Overall Rank (OR). Physical validity is evaluated at the initial scene configuration.
$\mathrm{Col}_{O}$ denotes the fraction of intersecting pairs among all reconstructed object pairs, excluding pairs involving the supporting table, while $\mathrm{Col}_{S}$ denotes the fraction of scenes containing at least one such inter-object intersection.

\subsection{Articulated-Object Evaluation Protocol}

We evaluate GRAM on all 243 assets from the Lightwheel benchmark~\citep{li2026particulate}, following its official evaluation protocol. Since GRAM operates on static object meshes, mesh-conditioned methods constitute the primary baselines, while image-conditioned methods are included as contextual references.

We report part-matching precision and recall, rest-pose mean Intersection over Union (mIoU), articulated generalized IoU (Art.\ gIoU), and whole-object Chamfer distance (OC). For the component ablation, we additionally report articulated-geometry point-cloud distance (Art.\ PC). Joint estimation is evaluated using angular error (AE) and axis-location error (LE). We also indicate whether each method requires articulation-specific training or fine-tuning.

\subsection{Manual Audit of Generated Articulations}
\label{app:generated_articulation_audit}

\paragraph{Evaluation setup.} We conduct an expert manual audit over the complete 150-scene evaluation set to assess articulation quality on the imperfect image-generated meshes encountered by the full reconstruction pipeline. A primary expert evaluates all candidate assets, and a second expert independently re-evaluates a uniformly random 20\% sample. Among the 150 scenes, 149 contain at least one candidate articulated asset, while one contains no movable object and therefore requires no joint-level assessment. The audit covers all 425 candidate assets and 527 predicted joints.

\paragraph{Rating protocol.} For each predicted joint, the evaluator inspects the input object crop, reconstructed initial state, and motion sweep, and separately scores structure and motion on a three-level scale. The structure score evaluates functional link grouping, joint type, axis, and pivot plausibility, while the motion score evaluates whether the predicted motion range remains functionally plausible without severe interpenetration, detachment, or geometric degeneration. A joint passes when both scores are 2, is partial when neither score is 0 and at least one score is 1, and fails when either score is 0. For multi-joint objects, the asset-level result is the worst result among its joints.

\begin{table}[t]
\centering
\small
\caption{\textbf{Manual audit of generated articulations over the complete 150-scene evaluation set.} Joint-level results include predicted joints only. Asset-level percentages are computed over the 418 judgeable assets; seven unjudgeable assets are excluded. Clear missing articulations are counted as asset-level failures.}
\label{tab:generated_articulation_audit}
\begin{tabular*}{\columnwidth}{@{\extracolsep{\fill}}lrrrr@{}}
\toprule
Evaluation level & $N$ & Pass & Partial & Fail \\
\midrule
Predicted joint & 527 & 454 (86.1\%) & 57 (10.8\%) & 16 (3.0\%) \\
Asset & 418 & 346 (82.8\%) & 55 (13.2\%) & 17 (4.1\%) \\
\bottomrule
\end{tabular*}
\end{table}

\paragraph{Results.} Among the 418 judgeable assets, 346 (82.8\%) pass, 55 (13.2\%) are partial, and 17 (4.1\%) fail. Among the 527 predicted joints, 454 (86.1\%) pass, 57 (10.8\%) are partial, and 16 (3.0\%) fail. Eleven candidate assets contain no predicted joint: four are judged as clear missing articulations and counted as asset-level failures, while seven lack sufficient visible evidence to determine whether an articulation is missing and are excluded as unjudgeable.

\paragraph{Inter-rater agreement.} To assess inter-rater reliability, the second expert independently re-evaluates 85 of the 425 candidate assets, corresponding to a uniformly random 20\% sample and 102 articulation evaluation items. As summarized in Table~\ref{tab:generated_articulation_agreement}, the two evaluators achieve 83.5\% exact agreement on the final asset-level rating, with Cohen's $\kappa=0.476$. Across the 101 items with paired numerical scores, exact agreement is 93.1\% for structure and 89.1\% for motion, with quadratically weighted $\kappa$ values of 0.474 and 0.659, respectively. Only one asset exhibits a direct pass--fail disagreement; the other disagreements occur between adjacent categories or involve unjudgeable versus fail.

\begin{table}[t]
\centering
\small
\caption{\textbf{Inter-rater agreement on the random 20\% re-evaluation sample.} The dagger denotes quadratically weighted Cohen's $\kappa$.}
\label{tab:generated_articulation_agreement}
\begin{tabular*}{\columnwidth}{@{\extracolsep{\fill}}lrrr@{}}
\toprule
Measure & $N$ & Exact agreement & Cohen's $\kappa$ \\
\midrule
Asset-level result & 85 & 83.5\% & 0.476 \\
Structure score & 101 & 93.1\% & 0.474$^\dagger$ \\
Motion score & 101 & 89.1\% & 0.659$^\dagger$ \\
\bottomrule
\end{tabular*}
\end{table}

\paragraph{Failure analysis.} We assign one primary reason to each of the 77 non-pass evaluation items. The identified causes are motion-range errors (17), part-segmentation errors (15), implausible axes or pivots (13), topology or joint-type errors (12), collision or detachment (10), and missing articulation (10).

\subsection{Implementation and Ablation Protocols}

We use GPT-5~\citep{singh2026openaigpt5card} for multimodal scene and object analysis, auxiliary-view generation, and the multimodal kinematic reasoning performed within GRAM. SAM 3~\citep{carion2026sam} provides class-agnostic instance segmentation, TRELLIS.2~\citep{Xiang_2026_Native} performs object-wise 3D reconstruction, and P3-SAM~\citep{ma2025p3sam} provides primitive segmentation for GRAM. VGGT~\citep{Wang_2025_VGGT} estimates dense point maps and camera parameters, while SASPS uses a frozen DINOv2 ViT-L/14 encoder. All pretrained components are used without task-specific training or fine-tuning.

Scene assembly and rendering are performed in Blender. Differentiable rasterization in PSGSR is implemented with nvdiffrast~\citep{laine2020modular}. Geometric collision checks within GRAM and scene-conditioned motion-range analysis use Trimesh/FCL~\citep{pan2012fcl}, while simulation-based physical evaluation is performed in Isaac Sim with PhysX.

All PSGSR variants share the same input images, proxy assets, and camera parameters, differing only in the evaluated registration components. GRAM component ablations follow a $2\times2$ factorial design over MLLM-guided joint fitting and semantic state reasoning. The marginal effect of each component is computed by averaging over the two settings of the other component.

Point-cloud alignment is optimized for up to $120$ iterations with a learning rate of $0.05$. In the full PSGSR pipeline, top-view and input-view refinement run for up to $100$ iterations at $512\times512$ and $1024\times1024$ resolution, respectively. The learning rates for translation, yaw, and scale are $0.03$, $0.03$, and $0.02$, respectively.

\subsection{Full GRAM Component Ablation}
\label{app:gram_ablation_full}

Table~\ref{tab:gram_ablation_full} reports the four component configurations underlying the averaged results in Table~\ref{tab:component_ablation}. The ablation is conducted as a separate evaluation run from the benchmark comparison in Table~\ref{tab:lightwheel}. Within the ablation run, all four configurations use the same 243 Lightwheel assets and official evaluation protocol, and all marginal effects are computed exclusively from these configurations.

\begin{table}[ht]
\centering
\small
\renewcommand{\arraystretch}{1.12}
\caption{\textbf{Full GRAM component ablation on Lightwheel.} Lower is better for all metrics.}
\label{tab:gram_ablation_full}
\begin{tabular*}{\linewidth}{@{\extracolsep{\fill}}lcccc@{}}
\toprule
Configuration
& AE ($^\circ$)$\downarrow$
& LE$\downarrow$
& Art.\ PC$\downarrow$
& OC$\downarrow$ \\
\midrule
Neither component
& 40.06 & 0.0527 & 0.2556 & 0.0177 \\
MLLM-Guided Joint Fitting only
& 17.17 & 0.0417 & 0.2653 & 0.0175 \\
Semantic State Reasoning only
& 40.06 & 0.0527 & 0.2552 & 0.0149 \\
Both components
& 17.17 & 0.0417 & 0.2452 & 0.0154 \\
\bottomrule
\end{tabular*}
\end{table}

\subsection{User Study Protocol}
\label{app:user_study_interface}
\begin{figure*}[t]
\centering
\includegraphics[width=\textwidth]{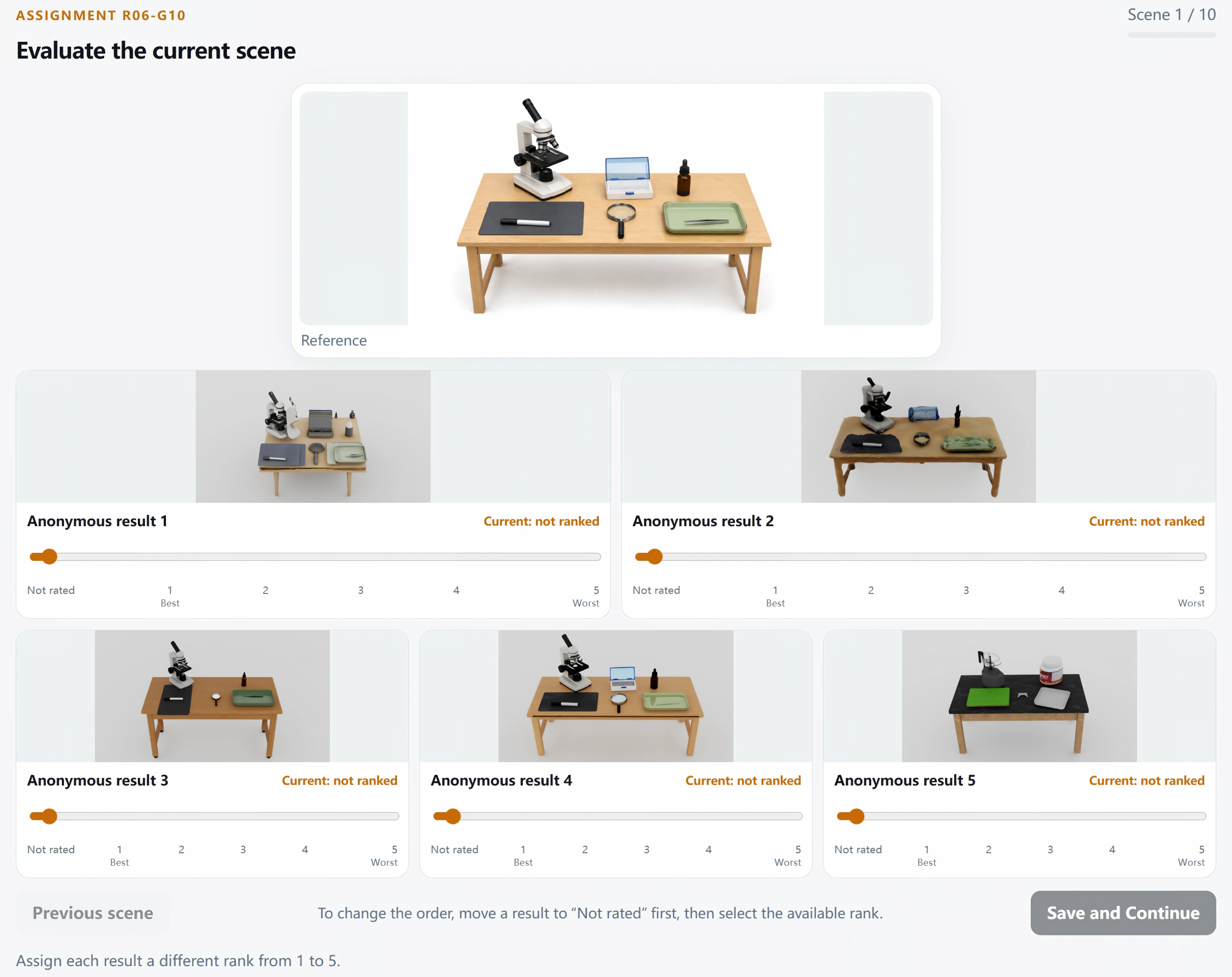}
\caption{\textbf{User-study interface.} The website interface used to present method outputs and collect participant rankings.}
\label{fig:user_study_interface}
\end{figure*}

Figure~\ref{fig:user_study_interface} shows the dedicated website used to present the outputs of ArticuTable and the four scene-reconstruction baselines for the same input scene. The study involved 59 adult volunteers, each of whom evaluated 10 scenes and ranked the five anonymized method outputs from 1 (best) to 5 (worst). Incomplete attempts were excluded, yielding 590 valid rankings per method. We report the mean rank, median rank, first-place rate, and top-two rate.

Among the participants, 72.9\% were aged 18--24 and 27.1\% were aged 25--34; 81.4\% identified as male and 18.6\% as female. All reported normal or corrected-to-normal vision and normal color vision. Regarding prior CG/3D vision experience, 55.9\% reported none, 22.0\% limited, 10.2\% moderate, and 11.9\% extensive experience. Participation was voluntary and uncompensated, and all participants provided informed consent before taking part. Only non-identifying demographic information was collected.

\section{Additional Qualitative Results}
\label{app:additional_qualitative_results}

\begin{figure*}[ht]
\centering
\includegraphics[
width=\textwidth,
height=0.88\textheight,
keepaspectratio
]{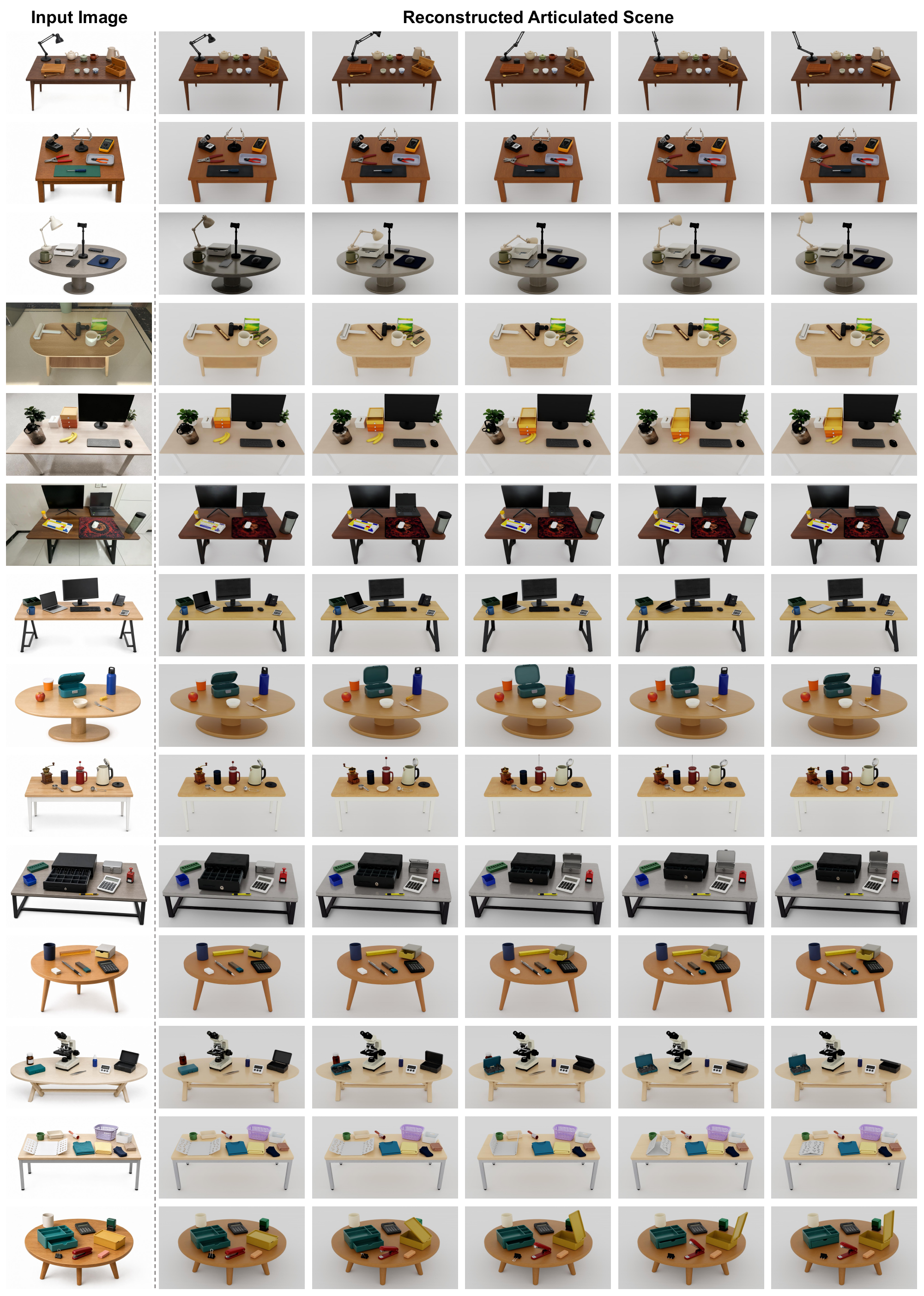}
\caption{\textbf{Additional qualitative results of articulated tabletop-scene reconstruction.} The first column shows the input images, while the remaining columns show the corresponding reconstructed scenes under different joint configurations.}
\label{fig:additional_articulated_scenes}
\end{figure*}

\clearpage

Figure~\ref{fig:additional_articulated_scenes} presents additional qualitative results of ArticuTable. For each input image, we render the reconstructed scene under multiple joint configurations. Across diverse tabletop layouts and object categories, ArticuTable preserves the identities and spatial arrangement of object instances while enabling coherent part-level articulation. These examples complement the quantitative evaluation by demonstrating the ability of ArticuTable to construct visually consistent and interactive articulated scenes.

\section{Runtime Analysis}
\label{app:runtime_analysis}

All runtimes are measured as per-scene wall-clock time on a single NVIDIA GeForce RTX 4090, excluding queueing time. Unless otherwise specified, the default parallelization strategy is enabled.

\paragraph{Stage-wise runtime.}

\begin{table}[ht]
\centering
\small
\caption{\textbf{PSGSR runtime.} Per-scene wall-clock time over 150 scenes, reported as mean $\pm$ standard deviation.}
\label{tab:psgsr_runtime_breakdown}
\renewcommand{\arraystretch}{1.1}
\begin{tabular*}{\columnwidth}{@{\extracolsep{\fill}}lr@{}}
\toprule
Stage & Time (s) $\downarrow$ \\
\midrule
Table and camera initialization & $8.3 \pm 2.4$ \\
Point-cloud alignment & $14.1 \pm 2.0$ \\
Top-view registration & $15.7 \pm 2.6$ \\
Input-view registration & $103.3 \pm 11.3$ \\
\midrule
Total PSGSR & $\mathbf{141.4 \pm 14.0}$ \\
\bottomrule
\end{tabular*}
\end{table}

Table~\ref{tab:psgsr_runtime_breakdown} reports the runtime of consecutive PSGSR stages over all 150 evaluation scenes. Input-view registration constitutes the dominant cost, accounting for approximately $73.1\%$ of the total runtime.

\paragraph{Cost of front-only pose selection.}

\begin{table}[ht]
\centering
\small
\caption{\textbf{Pose-selection efficiency.} Runtime averaged over the two scenes in Fig.~\ref{fig:qualitative_pose_selection_ablation}.}
\label{tab:front_only_runtime}
\renewcommand{\arraystretch}{1.1}
\begin{tabular*}{\columnwidth}{@{\extracolsep{\fill}}lcccc@{}}
\toprule
Variant & Initialization & Candidate Rendering & Time (s) $\downarrow$ & Relative \\
\midrule
Full PSGSR & Top-view-guided & Batched & \textbf{179.4} & $\mathbf{1.00\times}$ \\
Front-only SASPS & Input-view only & Batched & 374.8 & $2.09\times$ \\
Front-only SASPS & Input-view only & Sequential & 595.3 & $3.32\times$ \\
\bottomrule
\end{tabular*}
\end{table}

The top-view stage provides an efficient initialization by combining inexpensive 2D rotation search with SASPS. The selected pose restricts input-view SASPS to a few residual symmetry-related hypotheses; without this initialization, Front-only SASPS must instead render 3D candidates throughout the full yaw search. As shown in Fig.~\ref{fig:qualitative_pose_selection_ablation}, Front-only SASPS produces no obvious visual degradation relative to the full method. However, Table~\ref{tab:front_only_runtime} shows that this comparable result comes at a substantially higher computational cost: batched candidate rendering requires $2.09\times$ the runtime of the full method, while sequential candidate rendering further increases the runtime to $3.32\times$.

\section{Simulation-Ready Asset Validation}
\label{app:simready_validation}

This appendix details the export configuration and verification procedures underlying the simulation results in Sec.~\ref{sec:simulation_validation}. Package-integrity and initial-state checks cover all ArticuTable-100 scenes; editing and dynamic interaction are demonstrated in the representative scene shown in Fig.~\ref{fig:simready_validation}.

\subsection{USDZ Package Export and Configuration}
\label{app:usdz_export}

After scene registration and assembly, we convert each reconstructed scene into a self-contained USDZ package for direct use in Isaac Sim. The assembled rigid-object scene, including the reconstructed object poses, visual geometry, PBR materials, textures, and calibrated camera, is first exported from Blender as a USD visual layer. Each GRAM-generated URDF is subsequently imported into Isaac Sim as an independent articulation. Scene-level scale is baked into the corresponding link meshes and joint frames, while the reconstructed articulation-root pose and observed joint state are retained. The imported articulations are then composed with the rigid-object scene in a shared metric, Z-up world frame.

The visual and collision representations are kept separate throughout export. Visual meshes preserve their original geometry, materials, and textures. The supporting table uses its original unsimplified triangle mesh as an exact static collider. Ordinary movable rigid objects use high-fidelity convex decomposition with at most 128 convex hulls, at most 64 vertices per hull, a voxel resolution of $10^6$, a 1\% approximation error, and shrink wrapping enabled.

For articulated objects, the collision meshes stored in the intermediate URDF are not used as the final Isaac Sim collision representation. Instead, every articulated link receives an independent signed-distance-field (SDF) collider generated from its original link geometry. The SDF resolution and subgrid resolution are set to 128 and 6, respectively, using 16-bit subgrid values, a triangle-reduction factor of 0.25, and no SDF remeshing. Compared with a single convex hull, this per-link representation more faithfully preserves interaction-relevant concavities, including open containers, internal compartments, and through-holes.

The final stage contains a PhysX scene with gravity set to $9.81~\mathrm{m/s^2}$ along the negative Z-axis. Articulation bases remain unfixed so that complete articulated objects can be moved through physical interaction. Joint-drive stiffness and damping are both set to zero, allowing the joints to respond to external operation instead of being locked by position drives. The reconstructed root poses, object scales, observed joint positions, and intrinsic joint limits are authored directly into the USD stage. All saved rigid-body linear velocities, angular velocities, and joint velocities are initialized to zero. Dynamic collision geometries use a contact offset of 2 mm and a rest offset of 0.

Before packaging, all asset references are converted to paths relative to the root USD layer. Collision representations are constructed and checked in Isaac Sim, and the root layer stores \texttt{/physics/updateToUsd=true}. The root stage and all referenced USD layers, meshes, materials, and textures are then collected into a single archive using \texttt{UsdUtils.CreateNewUsdzPackage}. The resulting \texttt{isaac\_asset\_package.usdz} can therefore be transferred and opened without access to the original reconstruction directory. Table~\ref{tab:usdz_export_configuration} summarizes the export and physics configuration used for all released packages.

\begin{table}[t]
\centering
\small
\caption{\textbf{Configuration used to export the simulation-ready USDZ packages.}}
\label{tab:usdz_export_configuration}
\renewcommand{\arraystretch}{1.08}
\setlength{\tabcolsep}{4pt}
\begin{tabular}{@{}p{0.36\linewidth}p{0.58\linewidth}@{}}
\toprule
Setting & Configuration \\
\midrule
Package format & Self-contained USDZ \\
Stage units & $1$ unit $=1$ m \\
Up axis & $+Z$ \\
Gravity & $9.81~\mathrm{m/s^2}$ along $-Z$ \\
Visual representation & Original geometry with PBR materials and textures \\
Static table collision & Original unsimplified triangle mesh \\
Dynamic rigid collision & Convex decomposition \\
Maximum convex hulls & 128 \\
Maximum vertices per hull & 64 \\
Convex voxel resolution & $10^6$ \\
Convex approximation error & 1\% \\
Convex shrink wrapping & Enabled \\
Articulated-link collision & Per-link SDF from original link geometry \\
SDF resolution & 128 \\
SDF subgrid resolution & 6 \\
SDF subgrid precision & 16 bit \\
SDF triangle-reduction factor & 0.25 \\
SDF remeshing & Disabled \\
Contact / rest offset & 2 mm / 0 mm \\
Articulation base & Unfixed \\
Joint-drive stiffness / damping & 0 / 0 \\
Initial velocities & Zero \\
Physics-state update & \texttt{/physics/updateToUsd=true} \\
\bottomrule
\end{tabular}
\end{table}

\subsection{Direct-Package and Initial-State Verification}
\label{app:direct_package_verification}

To validate the released assets rather than intermediate pipeline states, we directly load each USDZ package into Isaac Sim and render its initial state with the simulation timeline stopped. We verify that the scene transforms and authored joint states remain unchanged during this zero-step rendering and that the package is not modified. We then reopen the same package, initialize the PhysX runtime, and restore its authored scene-entry state before applying gravity or external interaction. Thus, the observed settling, rigid-body motion, and articulation are produced by simulation rather than additional scene reconstruction or manual pose correction.

\subsection{Editing and Physics-Based Interaction}
\label{app:simready_interaction}

The editing examples in the top row of Fig.~\ref{fig:simready_validation} operate on independently addressable object instances. Removal deletes a selected instance, insertion adds a newly generated asset, and rearrangement changes existing instance transforms. Unchanged instances retain their reconstructed geometry and articulation without rerunning GRAM or PSGSR.

For the gravity-settling test in the bottom row, selected objects are initialized slightly above their target support regions before physics is enabled. Their subsequent settling is governed by gravity, collision, and contact. For the transport test, the object inside the toolbox has no rigid attachment or predefined parent--child connection to the toolbox; its motion during toolbox displacement arises from contact with the interior collision geometry. After displacement, the lid is opened through the recovered revolute joint while the contained object remains inside. This sequence tests support contact, preservation of the internal cavity, and articulation with per-link collision geometry in the assembled scene.


\end{document}